\documentclass[review]{elsarticle}

\usepackage{lineno,hyperref}
\usepackage{array}
\usepackage{chngcntr}
\usepackage{multirow}
\usepackage{threeparttable}
\usepackage{adjustbox}
\usepackage{comment}
\counterwithin{table}{section}
\counterwithin{figure}{section}
\modulolinenumbers[1]
\usepackage{fancyhdr}
\fancypagestyle{plain}{%
  \fancyhf{}
  \fancyfoot[C]{\iffloatpage{}{\thepage}}
  }
\usepackage{soul,xcolor}
\setstcolor{red}
\usepackage{rotating}

\journal{Renewable Energy}

\usepackage{numcompress}

\let\oldmaketitle\maketitle
\renewcommand{\maketitle}{\oldmaketitle\setcounter{footnote}{0}}

\begin{document}

\begin{frontmatter}

\title{Sky image-based solar forecasting using deep learning with multi-location data: training models locally, globally or via transfer learning?}

%
%
%

\author[stanford_affiliation]{Yuhao Nie\fnref{eqcontribute_note}\corref{correspon1}}
\author[cambridge_affiliation,engie_affiliation]{Quentin Paletta\fnref{eqcontribute_note}\corref{correspon2}}
\fntext[eqcontribute_note]{Equal contribution}
\author[stanford_affiliation]{Andea Scott}
\author[dewa_affiliation]{Luis Martin Pomares}
\author[engie_affiliation]{Guillaume Arbod}
\author[dewa_affiliation]{Sgouris Sgouridis}
\author[cambridge_affiliation]{Joan Lasenby}
\author[stanford_affiliation]{Adam Brandt}
\cortext[correspon1]{Corresponding author: ynie@stanford.edu}
\cortext[correspon2]{Corresponding author: qp208@cam.ac.uk}


\address[stanford_affiliation]{Department of Energy Resources Engineering, Stanford University, United States}
\address[cambridge_affiliation]{Department of Engineering, University of Cambridge, United Kingdom}
\address[dewa_affiliation]{Dubai Energy and Water Authority, Dubai, United Arab Emirates}
\address[engie_affiliation]{ENGIE Lab CRIGEN, France}

\begin{abstract}
Solar forecasting from ground-based sky images has shown great promise in reducing the uncertainty in solar power generation. With more and more sky image datasets open sourced in recent years, the development of accurate and reliable deep learning-based solar forecasting methods has seen a huge growth in potential. In this study, we explore three different training strategies for solar forecasting models by leveraging three heterogeneous datasets collected globally with different climate patterns. Specifically, we compare the performance of local models trained individually based on single datasets and global models trained jointly based on the fusion of multiple datasets, and further examine the knowledge transfer from pre-trained solar forecasting models to a new dataset of interest. The results suggest that the local models work well when deployed locally, but significant errors are observed when applied offsite. The global model can adapt well to individual locations at the cost of a potential increase in training efforts. Pre-training models on a large and diversified source dataset and transferring to a target dataset generally achieves superior performance over the other two strategies. With 80\% less training data, it can achieve comparable performance as the local baseline trained using the entire dataset.
\end{abstract}


\end{frontmatter}


\section{Introduction}
The continuous growth of solar photovoltaic (PV) deployment forms a critical part of the global energy transition. According to the International Energy Agency, a record-high 145 GW capacity has been installed during 2020 even in the face of the pandemic \cite{Masson2021}. The global cumulative PV capacity has amounted to 767 GW at the end of 2020, with around 70\% of the capacity installed over the last five years \cite{Masson2021}. The dramatic rise in PV installations will introduce challenges to the electricity grid due to the intermittency of solar energy, mainly caused by local and short-term cloud events \cite{Nie2021}. To reduce the uncertainty in solar power generation, accurate and reliable solar forecasting is thus pivotal and urgently needed.

There have been numerous research efforts toward building reliable solar forecasting models over the decades, which have targeted forecasting either the solar irradiance or the power output of PV systems. This introduction mainly focuses on the short-term or intra-hour solar forecasting research. Earlier efforts have tended to use statistical time series models \cite{Moreno-Munoz2008,REIKARD2009}, e.g., auto-regression (AR), auto-regression moving average (ARMA) and auto-regression integrated moving average (ARIMA), to auto-correlate irradiance/PV measurements for prediction. These methods usually lack forecast ability as they do not use any information on the movement and distribution of clouds \cite{Sun2019, yangReviewSolarForecasting2022a}. Since 2011, sky image-based solar forecasting has become more and more popular. Early works first extracted features from ground-based sky images, such as red-blue ratio, cloud coverage and cloud motion vectors, and then used these features for building physical deterministic models \cite{chow2011intra,marquez2013intra,quesada-ruizCloudtrackingMethodologyIntrahour2014a} or training machine learning models \cite{chuHybridIntrahourDNI2013a,Chu2015realtime,Chu2015reforcast,pedroAdaptiveImageFeatures2019}. In addition, several all-sky cameras can be used in stereo-vision mode to model the cloud cover in three dimensions to provide local irradiance maps \cite{peng3DCloudDetection2015, blancShorttermForecastingHigh2017a, kuhnValidationAllskyImager2018a}. In the past five years, with the further development of computer vision techniques, efforts have shifted to concentrate on building end-to-end deep learning models, such as convolutional neural networks (CNNs) \cite{Sun2018,Sun2019,Venugopal2019,Nie2020,Feng2020, palettaConvolutionalNeuralNetworks2020,Nie2021,Feng2022} or CNNs hybridized with recurrent neural networks (RNNs) \cite{Zhang2018,palettaBenchmarkingDeepLearning2021,Paletta2021eclipse} to correlate irradiance/PV with sky images, which have generally achieved superior performance over the other methods despite some limitations \cite{palettaBenchmarkingDeepLearning2021}. These modern methods will be the focus of this study.

One of the most important factors for deep learning-based solar forecasting models is the dataset~\cite{nieOpenSourceGroundbasedSky2022}. To train a generalized deep learning model that works well not only on the model development dataset but also for unseen data, the training dataset needs to contain massive amounts of imagery and diversified samples with numerous sky conditions. Contrary to simulated datasets which can be easily extended, real-world data such as sky images are constrained by the period of collection. For this reason, researchers in the solar forecasting community have explored using data augmentation, a common technique in deep learning, to artificially increase the diversity of image samples \cite{Nie2021, palettaSPINSimplifyingPolar2021}. Although data augmentation is viable without accessing additional data, the benefit it could bring is constrained by the diversity of cloud patterns in the dataset. With more and more sky image datasets open-sourced in recent years \cite{kurtzVirtualSkyImager2017, pedroComprehensiveDatasetAccelerated2019, fengOpenSolarPromotingOpenness2019, terren-serranoGirasolSkyImaging2021, ntavelisSkyCamDatasetSky2021, nie2022skippd}, potential options have not been explored thoroughly in solar forecasting, including training a model by fusing different datasets with more or less heterogeneity and knowledge sharing between different datasets via pre-trained models. The primary objective of this study is to investigate multi-location augmentation approaches that could be implemented by other users to take advantage of open source datasets for their specific computer vision tasks (e.g., solar forecasting, cloud masking, cloud motion prediction, radiative transfer modelling).

Specifically, we examine the following questions by using datasets collected globally from three different locations with disparate climate patterns (the details on these three datasets can be found in Section \ref{sec:dataset}): 
\begin{enumerate}
    \item Should deep learning models be trained locally using the location-specific dataset or globally via fusion of datasets from different locations?
    \item How to deal with the dataset heterogeneity for training global models, especially the different scales and distributions of prediction targets?
    \item Is there any knowledge that can be shared between different locations via transfer learning from pre-trained models?
\end{enumerate}

To address these questions, two research groups from Stanford University and Cambridge University, as well as researchers from the Dubai Energy and Water Authority (DEWA), have worked in parallel and developed deep learning models for a short-term solar forecasting task, the goal of which is to predict 15-min-ahead PV power outputs (or irradiance values) based on the imagery and measurement data collected in the past 15 minutes (see Section \ref{sec:methodology} for more details on the model setup). Two different deep learning model architectures are utilized by the two University teams to avoid bias caused by particular model architectures and ensure the reliability of the results for understanding the impact of diverse training data.

The rest of this paper is organized as follows: in Section \ref{sec:literature review}, we
review the methods for training deep learning models with multi-location data, including dataset fusion and transfer learning. Section \ref{sec:dataset} describes the sky images and PV output/irradiance datasets used in this study from three different locations around the world. Section \ref{sec:methodology} presents the methodology, including the model architectures, training details and evaluation metrics. Section \ref{sec:experiments} delves into the experimental designs for exploring the optimal strategies for training PV/irradiance prediction models with multi-location data. Section \ref{sec:results and discussion} analyzes and discusses the results and provides directions for future research. Finally, we summarize the findings of this study in Section \ref{sec:conclusion}.

\section{Review of multi-location dataset modeling}
\label{sec:literature review}
Two methods for multi-location dataset modeling are reviewed in this section. The first type is dataset fusion, basically integrating the datasets from multiple locations and training the model jointly. This approach is based on the expectation that fused data are more informative than any individual dataset. Another method which is widely used in the deep learning community but not yet well studied in image-based solar forecasting, is transfer learning. In transfer learning, the datasets are used sequentially, with model parameters passed from earlier to later models to pass learning forward in the training process.

\subsection{Dataset fusion}
Dataset fusion is commonly used in image-based solar forecasting studies with multi-location datasets. \citet{Pothineni2019} experimented with two sky image datasets collected in two regions with identical camera setup: one in Italy and one in the Swiss mountains. Irradiance measurements were used to determine the associated sky conditions to be either clear or occluded by thresholding them with the irradiance values derived from a clear sky model. The authors compared training CNN models on the individual datasets with training on the fused dataset for predicting the 5-min ahead sky conditions, with results showing superior performance from training the model jointly on both datasets. \citet{Bansal2021} developed a CNN-LSTM auto-regressive model to predict the satellite spectral channel values at the target sites based on a sequence of past satellite observations. They combined satellite data across 25 solar sites in the US to train one global model and test it on individual locations. The authors suggested that while training location-specific models could identify unique local attributes to improve accuracy, their deep learning model is quite data hungry. Under the situation where there is not enough data to train and test on any single location, a global model might be beneficial as it can be applied to any location without re-training. 

One point to note here is that most of the existing studies using dataset fusion methods are based on the premise that the prediction targets are of similar scales, e.g., solar irradiance and satellite spectral channel value mentioned in studies above. Very few projects have studied fusing datasets with different scales or different units of the prediction targets, e.g., PV output measurements from systems with different capacity, or PV output and irradiance measurements which are highly correlated. Another challenge that needs attention is that the imagery data collected by these studies are generally based on similar camera setups, e.g., camera model and placement orientation. It is possible that in future fusion studies multiple data streams would be generated by different camera setups (e.g., resolution, contrast, color balance, spectral range, orientation).

\subsection{Transfer learning} 
Transfer learning aims at improving the performance of target learners or solving new problems faster on target domains by transferring the knowledge learned from different (but related) source domains \cite{zhuangComprehensiveSurveyTransfer2021}. A condition for the transfer of knowledge is the existence of a similarity between tasks. In solar energy for instance, a solar site could benefit from datasets generated in other locations to improve the accuracy of its site-specific algorithm. This specific transfer learning approach, termed domain adaptation, aims at adapting a learning algorithm to a new data distribution while keeping the task unchanged. This could be especially beneficial for new solar facilities which have a limited amount of data. Other similar but distinct activities in solar forecasting are, for instance, PV power output \textit{versus} solar irradiance forecasting, or cloud cover modelling from sky images \textit{versus} satellite observations.

There are different ways to categorize transfer learning approaches. An approach is known as \textit{homogeneous transfer learning} if the type of input variables (e.g., sky images) and labels (e.g, irradiance) are the same for source and target domains (location A and B). In contrast, if input variables (e.g., sky images versus satellite observations) or labels (e.g., irradiance versus PV output) are distinct, an approach is referred to as \textit{heterogeneous transfer learning} \cite{zhuangComprehensiveSurveyTransfer2021}. Another review splits methods into four groups: instance-based (instance weighting strategy), feature-based (creation of a new feature representation), parameter-based (the transferred knowledge is encoded at the parameter/model level) and relational-based (transfer the relationship among the source data to the target domain) approaches \cite{panSurveyTransferLearning2010}. Instead of learning a new task from scratch, a common practice in deep learning is to start learning a new task such as solar forecasting~\cite{wenDeepLearningBased2021} with standard neural networks  VGGNet~\cite{simonyanVeryDeepConvolutional2015}, ResNet~\cite{heDeepResidualLearning2016}, DenseNet~\cite{huangDenselyConnectedConvolutional2017}, etc., pretrained on large datasets, e.g. ImageNet~\cite{dengImageNetLargescaleHierarchical2009}. Alternatively, self-supervised learning, a form of unsupervised learning that generates pseudo-labels from the data itself, offers a framework to pretrain a model on an unlabeled dataset close to the target dataset instead of on a generic dataset. We describe in the following paragraph some studies which have applied transfer learning techniques to image-based solar forecasting and related fields.


In sky image segmentation, a data representation can be learnt on a large unlabelled dataset of sky images with tasks such as image reconstruction, clustering and classification, prior to fine-tuning the model on the target segmentation task with a smaller labeled dataset \cite{fabelApplyingSelfsupervisedLearning2021}. Models pretrained on the sky image dataset outperform those pretrained on ImageNet  \cite{dengImageNetLargescaleHierarchical2009} or randomly initialised. \citet{Pothineni2019} claimed the deep learning model they developed (KloudNet) can be used to improve the performance on other PV plants via transfer learning, but no quantitative results are presented in their study. In solar irradiance estimation from satellite data, a recent work applied transfer learning to a four-layer neural network \cite{liEstimatingGlobalDownward2022}. The source domain consists of simulated data used to pre-train the model while real-world satellite and in-situ observations are used for fine-tuning the last four layers. Although some negative transfer is observed, the proposed transfer learning approach benefits some tasks such as daily downward shortwave radiation estimation compared to baselines trained on simulated or in-situ data only.

\section{Dataset}
\label{sec:dataset}
\subsection{Dataset overview}

In our study, three datasets collected globally from three different locations with drastically different weather conditions are used for experiments in this study. These datasets are (1) the Stanford dataset~\cite{nie2022skippd}, collected on the campus of Stanford University in California, United States (US), which is characterized by long summers with mostly clear sky and short winters with partly cloudy sky; (2) the SIRTA dataset \cite{sirta}, collected by the SIRTA Atmospheric Observatory in Palaiseau, France, which is dominated by partly cloudy and cloudy sky conditions over much of the year, and (3) the DEWA dataset, collected from the outdoor testing facility of the Dubai Energy and Water Authority (DEWA) in the United Arab Emirates (UAE), which is clear most of the time over a year but with sandstorms occurring usually in dry summers. Among the three datasets, the Stanford and SIRTA datasets are open source and can be accessed by the public \footnote{Detailed information about the Stanford dataset can be found via the following Github Repository \url{https://github.com/yuhao-nie/Stanford-solar-forecasting-dataset}; the SIRTA dataset is available upon request to \url{https://sirta.ipsl.polytechnique.fr/data_form.html}}, while the DEWA dataset is private and not publicly available. It should be noted that only team Stanford has access to the DEWA dataset due to an internal collaboration contract between Stanford and DEWA, and team Cambridge only has access to the two publicly available datasets. The detailed comparison of these datasets is shown in Table \ref{tab:dataset comparison}.

\vspace{0.5\baselineskip}
\begin{table}[h!]
\begin{center}
\caption{Comparison of three studied datasets}
\begin{small}
\begin{tabular}{>{\raggedright}p{0.25\linewidth}>{\raggedright}p{0.21\linewidth}>{\raggedright}p{0.21\linewidth}>{\raggedright\arraybackslash}p{0.21\linewidth}}
\hline
 & Stanford & SIRTA & DEWA \\
\hline\hline
\noalign{\vskip 1mm}
Location & Stanford, US & Palaiseau, France & Dubai, UAE \\
Data type & sky images \& PV power output & sky images \& global horizontal irradiance (GHI) & sky images \& GHI \\
Data frequency & 1-min & 1 to 2-min$^*$ & 1-min\\
Image resolution$^{**}$ & $2048^2\rightarrow64^2$  & $650^2\rightarrow64^2$ & $1920^2\rightarrow64^2$ \\
Time window & 2017.3--2019.11 & 2017.1--2019.12 & 2021.1--2021.11\\
Camera model & Hikvision DS-2CD6362F-IV & EKO SRF-02 & EKO ASI-16 \\
Camera orientation & 14\textdegree\ south by west & 2\textdegree\ south by east & due south \\
PV system & 30-kW rooftop system with elevation 22.5\textdegree, azimuth 195\textdegree & N/A & N/A \\
Number of days & 269 & 522 & 953\\
Number of valid samples & 135,527 & 448,268 & 91,979 \\
Training, validation and test set split & 84\%:9\%:7\% & 88\%:10\%:2\% & 83\%:9\%:7\% \\
\noalign{\vskip 1mm}
\hline
\end{tabular}
\end{small}
\end{center}
\label{tab:dataset comparison}
\footnotesize{$^*$ For the image data, the frequency is 1-min in 2017 and 2-min in 2018 to 2019; for the irradiance data, the frequency is 1-min for 2017 to 2019; $^{**}$ $\rightarrow$ means image down-sizing from the high resolution raw images}
\end{table}

\begin{figure}
\centering    
\includegraphics[width=0.8\textwidth]{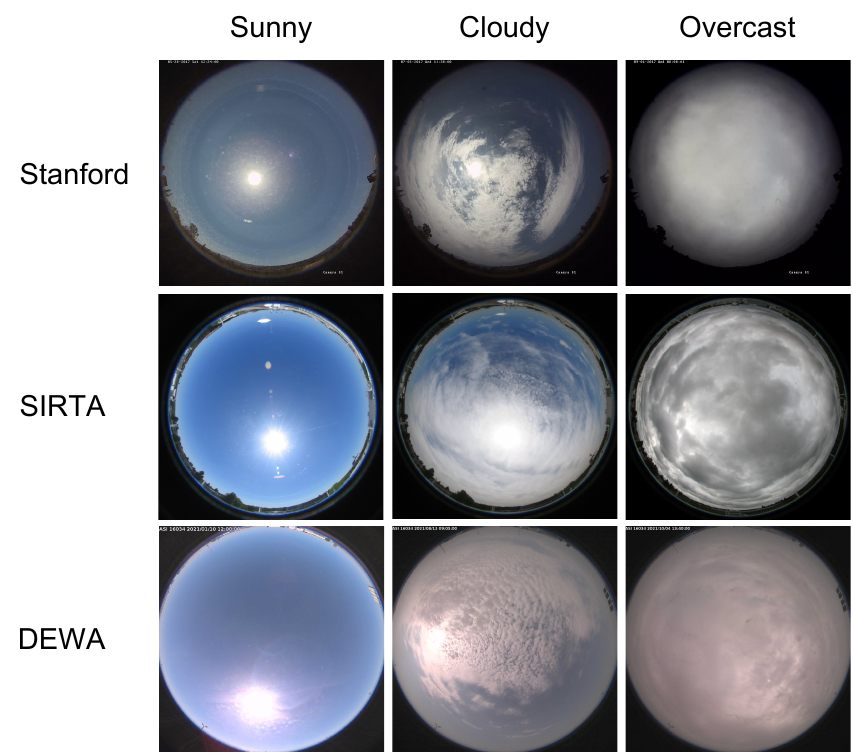}
\caption{Sky image examples from three locations with different weather conditions}
\label{fig:sky_image_samples}
\end{figure}

The different weather conditions can be reflected by the data from these three locations. Figure \ref{fig:sky_image_samples} shows sky image examples from three locations with different weather conditions, sunny, cloudy and overcast, respectively. It can be observed that the images from the three datasets look somehow different especially the hue, for example, DEWA image samples appear yellowish and dusty due to the impact of sandstorms. In addition, different camera characteristics, e.g., camera contrast and saturation, can affect the images.

\subsection{Data processing}
To ensure consistent data processing, team Stanford processed the data from the three datasets and then shared with team Cambridge for all the experiments. The high-resolution raw images collected by the cameras are first down-sized to $64\times64$ pixels to save computation in the model training process. The PV output and irradiance measurements collected from the data loggers are averaged over a minute. To form valid samples for the forecasting task, each time point ($t_0$) is checked to ensure the availability of (1) the prediction target, i.e. the PV output/GHI measurements 15 minutes ahead ($t_0+15$); and (2) the model inputs, i.e. the sky images and concurrent PV output/irradiance measurements over the past 15 minutes at a 2 minutes resolution ($t_0-14, t_0-12, ..., t_0$). All time stamps that do not satisfy these two conditions are dropped. For the Stanford dataset and the DEWA dataset, the sampling frequency is chosen to be 2 minutes (two samples differ in $t_0$ by 2 minutes) because higher frequency led to a longer model training time with limited improvement on the model accuracy (\cite{Sun2019}). For SIRTA dataset, the sampling frequency is set to be 1 minute for 2017 and 2 minutes for 2018 to 2019 to be consistent with the imagery data frequency (see Table \ref{tab:dataset comparison}). After processing, the number of valid samples for the three datasets can be found in Table \ref{tab:dataset comparison}. In this study, the three datasets share similar characteristics such as similar camera orientations and their location in the same hemisphere. However, these aspects might partially hinder the transfer of knowledge in other contexts (e.g., different trajectory of the sun due to the location of the camera or its orientation). This could be addressed by the application of data augmentation techniques in the data processing stage. For instance, a transfer function independent of the camera orientation can be learnt by randomly rotating sky images during training or by representing the scene with polar coordinates centered on the sun~\cite{palettaTemporallyConsistentImagebased2020, palettaSPINSimplifyingPolar2021, palettaCLoudFlowCentring2022}, which will be explored in the future by accessing more diversified datasets around the world.

\subsection{Dataset partition}
\begin{figure}
\centering    
\makebox[\textwidth][c]{\includegraphics[width=1.1\textwidth]{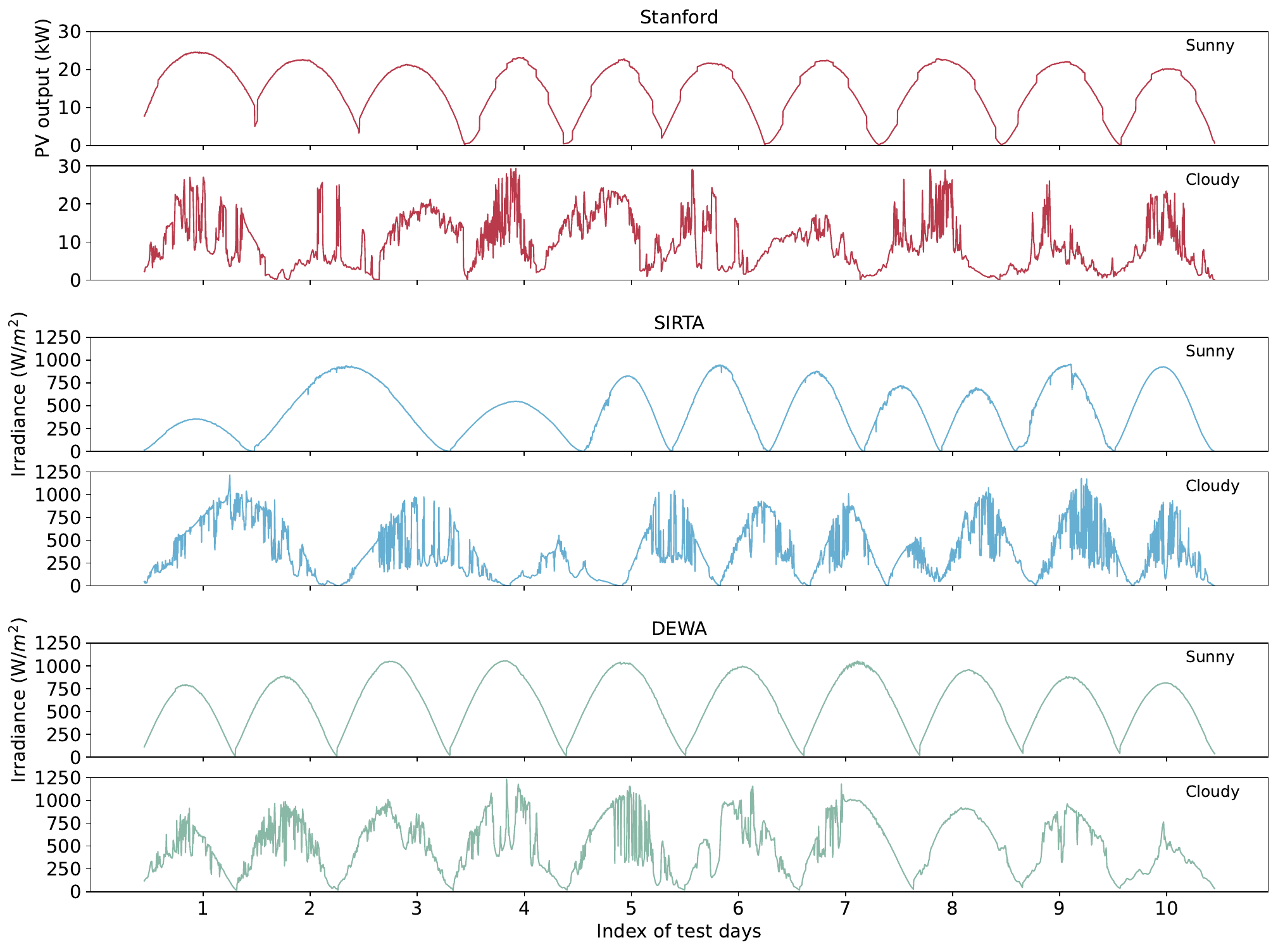}}
\caption{The test set PV/irradiance profiles for the three datasets (The zigzags in the sunny days of the Stanford test set is due to missing data points)}
\label{fig:test_set_profile}
\end{figure}

\begin{figure}
\centering    
\makebox[\textwidth][c]{\includegraphics[width=1.1\textwidth]{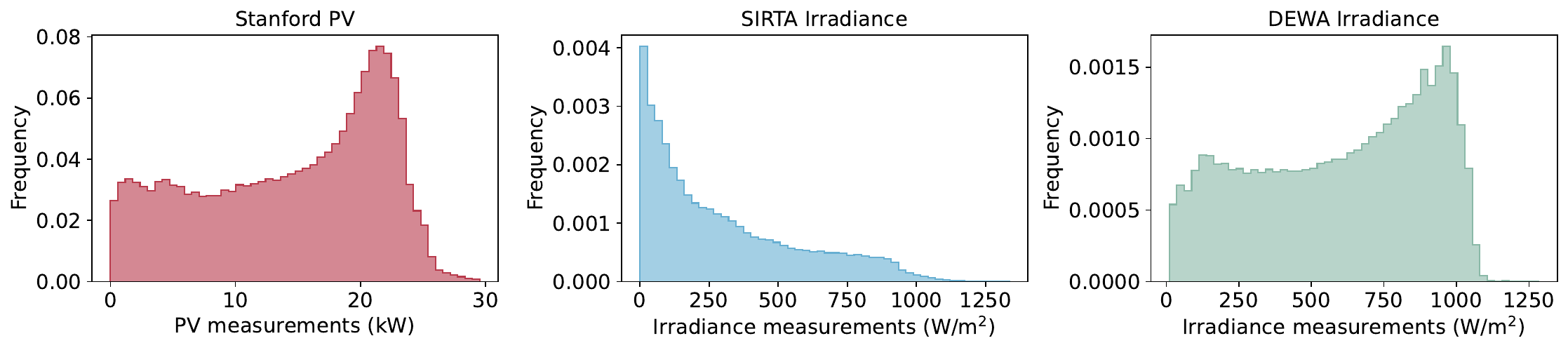}}
\caption{The model development set data distribution of the three locations} 
\label{fig:pv_irradiance_distribution}
\end{figure}

For model development and evaluation purposes, the valid samples of all three datasets are partitioned into the model development set (consisting of training and validation) and the test set. The test set is first separated out with 10 sunny days and 10 cloudy days across the entire time period and is never touched during the model development processes. The PV output/irradiance profiles of these 20 days for all three datasets are shown in Figure \ref{fig:test_set_profile}. The remaining data go to the model development set. Figure \ref{fig:pv_irradiance_distribution} presents the model development set data distribution of the three locations. It can be observed that Stanford PV and DEWA irradiance distributions share some similarity while SIRTA irradiance distribution shows a nearly opposite trend compared to the other two datasets. To avoid the bias from data partitioning and obtain a less optimistic estimate of the model performance, ten-fold cross-validation is employed in this study, which divides the development set into 10 folds, 9 folds for training the model and 1 fold for validating the model. The model is trained 10 times, each time with a different fold as the validation set, resulting in 10 sub-models. The final prediction made by the model is the ensemble mean of these 10 sub-models. Under this setup, the split of training, validation and testing samples in percent for each dataset is shown in Table \ref{tab:dataset comparison}. Moreover, to avoid the data from the same day ending up in both training and validation sets, which could potentially lead to over-estimates of the model performance due to the closeness between training and validation samples, day-block shuffling is performed during the cross-validation process \cite{Sun2019}.

\section{Methodology}
\label{sec:methodology}
The specific solar forecasting task tackled in this study is to predict 15-min-ahead PV power output or GHI values based on the three datasets described in Section \ref{sec:dataset}. Two different deep learning models utilized by the two teams are presented in this section, including the model architectures and training details, meanwhile, the metrics used to evaluate the performance of the models are also described. It should be noted that, we do not focus on developing new solar forecasting methods in this study, rather we mainly focus on comparing the effectiveness of different model training strategies based on published solar forecasting models, namely, SUNSET \cite{Sun2019} and ConvLSTM \cite{palettaBenchmarkingDeepLearning2021}, which have already been compared with existing methods in the published articles. We thus refer the readers to these two published articles for more details in terms of methodology comparison with existing methods. Meanwhile, to avoid the bias caused by particular model architectures and ensure the reliability of the results for understanding the impact of diverse training data, we have conducted most of the experiments in this study using both model architectures mentioned above.

\subsection{Deep learning models}
\label{subsec:deep learning models}
\begin{figure}
\makebox[\textwidth][c]{\includegraphics[width=1.2\textwidth]{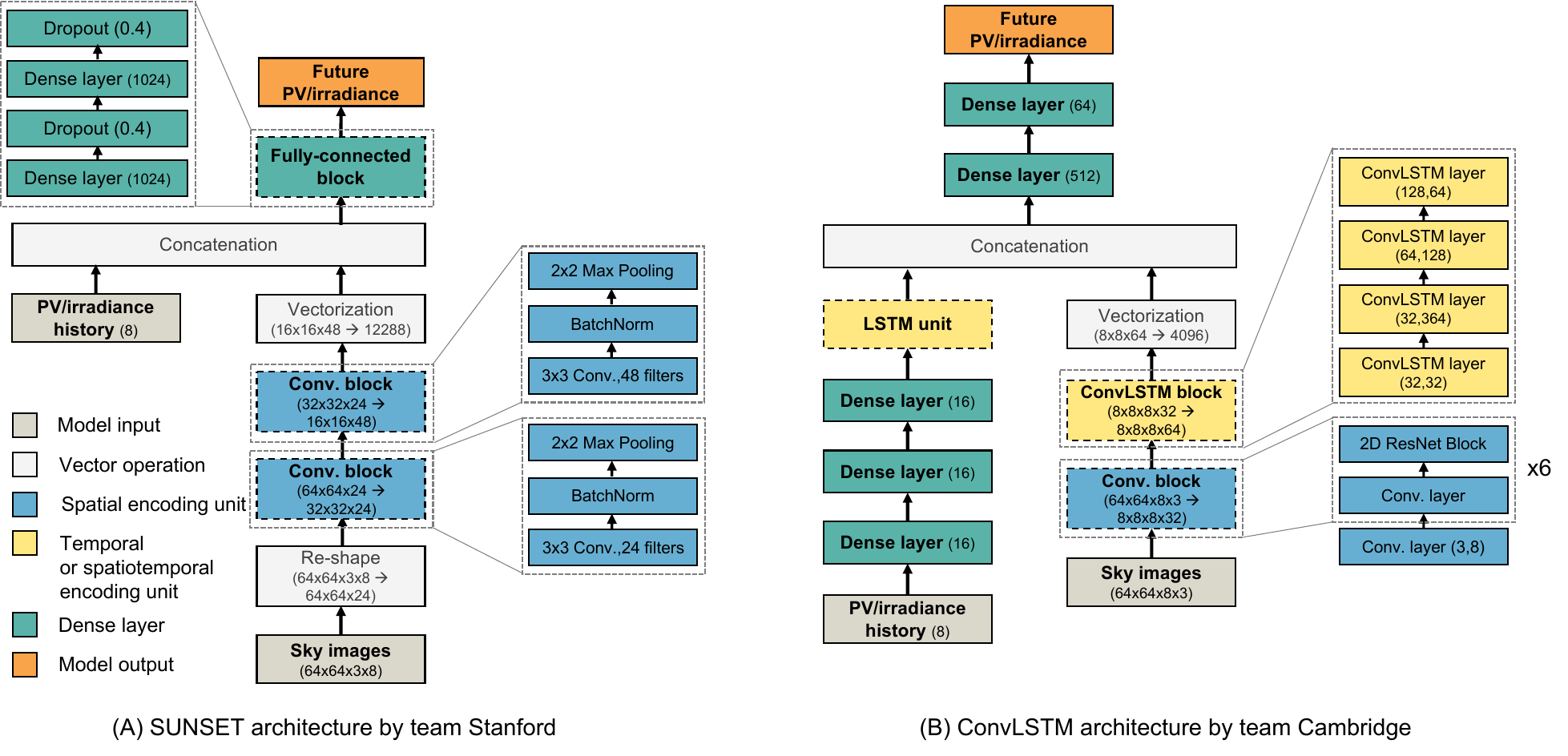}}
\caption{Model architectures used by team Stanford and Cambridge.}
\label{fig:model_architecture}
\end{figure}

While the two teams utilized different deep learning models for the forecasting task, the following common setups are shared: (1) using hybrid features as the model input --- sky images and measurements history (PV output/GHI) in the past 15 minutes with 2-min interval; (2) minimizing the mean squared error (MSE) loss function and using the stochastic gradient descent optimizer Adam \cite{Kingma2014} for model training; (3) employing 10-fold cross-validation and training 10 sub-models; and (4) for model evaluation, the prediction is generated by the ensemble mean of the predictions from the 10 sub-models. An illustration of the model architectures of both teams can be found in Figure \ref{fig:model_architecture} and the comparisons of the architectures as well as training settings are listed in Table \ref{tab:model comparison} with details described individually in the subsections below.

\begin{table}[h!]
\begin{center}
\caption{Comparison of the two deep learning architectures and training details}
\begin{small}
\begin{tabular}{>{\raggedright}p{0.40\linewidth}>{\raggedright}p{0.3\linewidth}>{\raggedright\arraybackslash}p{0.2\linewidth}}
\hline
 & SUNSET & ConvLSTM \\
\hline\hline
\noalign{\vskip 1mm}
Input sky images & Past 8 frames & Past 8 frames \\
Input PV or GHI history & Past 8 values & Past 8 values \\
Activation functions & ReLU & ReLU \\
Normalization layers & Batch normalization & None \\
Pooling layers & Max pooling & None \\
Dropout rate & 0.4 & 0.0 \\
Number of Parameters & 13.66M & 4.25M \\
- Number of convolutional layers & 2 & 7 \\
- Number of residual blocks & 0 & 6 \\
Optimizer & Adam & Adam \\
Training loss & MSE & MSE \\
Learning rate & $3 \times 10^{-6*}$/$2.5\times 10^{-5**}$ & $1 \times 10^{-5}$ \\
Batch size & 256 & 10 \\
\noalign{\vskip 1mm}
\hline
\end{tabular}
\end{small}
\end{center}
\label{tab:model comparison}
\footnotesize{$^*$ for training on Stanford dataset; $^{**}$ for training on SIRTA dataset}
\end{table}

\subsubsection{Stanford SUNSET model} 
A CNN architecture named SUNSET (Stanford University Neural Network for Solar Electricity Trend) is used by team Stanford. The SUNSET model is first introduced by \citet{Sun2019} to forecast 15-min-ahead PV output and is characterized by its usage of CNN and hybrid input features. The 1-min lag term interval used in the original SUNSET model is changed to 2-min in this study and modifications were made accordingly to accommodate the different input and output for each dataset, either PV or irradiance measurements. 

The basic structure of the SUNSET model includes two Convolutional (Conv.) blocks and one Fully-connected (FC) block. The Conv. block employs a sandwich-like structure, including sequentially one Conv. layer, one batch normalization (BatchNorm) layer, and one pooling layer. The Conv. layer utilizes a 3×3 filter, with a stride of 1 and same-value padding. The activation function used for the Conv. layer is a rectified linear unit (ReLU). In the pooling layer, 2×2 max pooling with a stride of 2 is used to reduce the activation spatial dimensions. The first Conv. block contains 24 filters, while the second contains 48 filters. After the two Conv. blocks, the processed input is vectorized and concatenated with PV output/irradiance history and passed through the FC block to produce the prediction. The FC block includes two dense layers, each containing 1024 neurons and using ReLU as its activation function. After each dense layer, a dropout layer with a 0.4 dropout rate is performed to prevent over-fitting. 

Different learning rates are used for model training on different datasets by team Stanford. For the Stanford dataset, a learning rate of  $3 \times 10^{-6}$ is used as this was seen to be effective via prior studies \cite{Sun2019}, while for the SIRTA dataset, a learning rate of 2.5$\times 10^{-5}$ is used based on initial hyper-parameter selection experiments. Batch size is consistently set to be 256 for the stochastic gradient descent optimizer. An early stopping scheme is applied to prevent potential over-fitting, namely, the training is stopped when the validation loss is not observed to decrease for five consecutive epochs.

\subsubsection{Cambridge ConvLSTM model}
A ConvLSTM model architecture is used by team Cambridge, which was first presented by \citet{palettaBenchmarkingDeepLearning2021} to forecast 10-min ahead GHI levels from past sky images and auxiliary data (irradiance measurements and solar position). Several modifications were made to the original model architecture to this study. The first convolutional layer with a stride of two was changed to a layer with unitary stride to account for the lower spatial resolution of images in the present study (64 pixels instead of 128). In addition, the model now takes three-channel images (RGB) as input instead of two (grey-scale short and long exposure images). Furthermore, the new model is fed only with the past GHI or PV output measurements as auxiliary input.

The ConvLSTM model is made of two parallel encoders for auxiliary data and sky images. Features from past irradiance or PV measurements are extracted with three dense layers and an LSTM layer \cite{Hochreiter1997}. Sky images are filtered through a sequence of convolutional layers and residual blocks \cite{heDeepResidualLearning2015} to decrease the spatial resolution from $64 \times 64$ to $8 \times 8$. The resulting set of feature maps is then sequentially fed into a ConvLSTM module \cite{Shi2015}. A learning rate of $1 \times 10^{-5}$ is used to train the ConvLSTM model for both datasets with a batch size of 10.

\subsection{Evaluation metrics}
\label{subsec:eval_metrics}
In this study, we evaluate the model performance from two aspects: (1) the prediction accuracy, which is measured by some common error metrics via applying the trained models to the test sets; and (2) the training cost, which reflects the consumption of computational resource during the model training.

To assess the prediction accuracy, the error metric root mean squared error (RMSE) is used. RMSE is the most commonly used metric and can be expressed by Equation (\ref{eq:RMSE}); other similar metrics like mean absolute error (MAE), mean bias error (MBE), etc. are not covered in this study.
\begin{equation}
    \mathrm{RMSE} = \sqrt{\frac{1}{N}\sum_{i=1}^N(\hat{Y}_i - Y_i)^2}
    \label{eq:RMSE}
\end{equation}
where $N$ is the number of samples, $\hat{Y}$ is the prediction generated by the model and $Y$ is the ground-truth measurement.

For evaluating the training cost, we define a metric called training effort (TE), which is essentially the total number of samples seen by the model until convergence and can be expressed by Equation (\ref{eq:training_effort}):
\begin{equation}
    \mathrm{TE} = \mathrm{Number\ of\ training\ epochs}\times \mathrm{Training\ set\ size}
    \label{eq:training_effort}
\end{equation}

The number of epochs is obtained when the model stops training. Specifically, team Stanford uses an early stopping scheme for the SUNSET model, and the training is stopped when the validation loss does not decrease for five consecutive epochs. Team Cambridge stops the training when the ConvLSTM model starts overfitting, i.e. when the validation loss increases for several consecutive epochs. We do not compare the training time in this study, as it varies with the different GPU models used in training (Stanford used Tesla A100, while Cambridge used GeForce GTX 1080).

 It should be noted that in previous publications \cite{Sun2019,palettaBenchmarkingDeepLearning2021}, the proposed models, namely, SUNSET and ConvLSTM, have already been compared with reference models such as auto-regressive model and smart persistence model and evaluated with various metrics such as the forecast skill, the ramp score, temporal distortion metrics and scatter plots \cite{Vallance2017}. Given the scope of this study, we thus did not conduct such comparison and evaluation for the models here.
 
\section{Experiments}
\label{sec:experiments}
This section details the experiments conducted to address the questions raised in the Introduction section. First, the local models are trained individually based on the dataset from each location, which serve as the baselines for comparison with other models developed in this study. Next, a global model is trained jointly based on the combination of datasets from different locations. In view of the sample heterogeneity of different datasets, we investigated the effect of different input and output normalization methods on the model performance and modifications of the model architectures to accommodate input data streams from different locations. Moreover, we explore the potential of knowledge transfer between different datasets via transfer learning.

\subsection{Training local models}
\label{subsec:training_local_models}
The two research teams first train their models individually on each one of the three datasets to construct a local baseline. To examine the generalization of the local models, in other words, to test how well the local models perform when applied offsite without re-training, the models are trained based on the normalized data instead of the original data. Training on normalized data enables the models to be deployed across different locations with different scales or types (e.g., trained on PV data but applied to predict irradiance values) of the prediction targets. Specifically, during the development phase, the models learn to predict the relative values of PV power output or irradiance, and during the implementation phase, the relative predictions generated by the models are post-processed to revert them to the original scale. Here, the data normalization method used by each team is the optimal normalization method identified when the training model jointly uses integrated datasets. Details can be found in Sections \ref{subsec:training_global_models} and \ref{subsec:global_model_results}.

\subsection{Training global models} 
\label{subsec:training_global_models}
The challenges of training global models are associated with the heterogeneity of samples from different datasets. A common problem is the different scales (e.g., 30-kW PV system versus 2-MW PV system) and/or different types (e.g., irradiance versus PV output) of the measurement data. In this study, we mainly deal with the latter. Moreover, differences in the camera setups or data distribution caused by the unique local weather conditions, could add other layers of difficulty. The PV output data in the Stanford dataset and the irradiance measurements in the SIRTA dataset are a good example demonstrating all of the above issues (see figure \ref{fig:pv_irradiance_distribution}), and thus will be our main focus in this section. Other possible combinations of datasets (e.g., Stanford+DEWA, Stanford+SIRTA+DEWA) could essentially go through similar training processes. Specifically, to combine the two datasets for joint training, we explored using different normalization methods for processing the model input and output data. We also experimented with tuning the model architectures to better accommodate the multiple input data streams.

\subsubsection{Input and output normalization}
\label{subsubsec:input and output normalization}
\begin{figure}[h!]
\centering    
\includegraphics[width=1.0\textwidth]{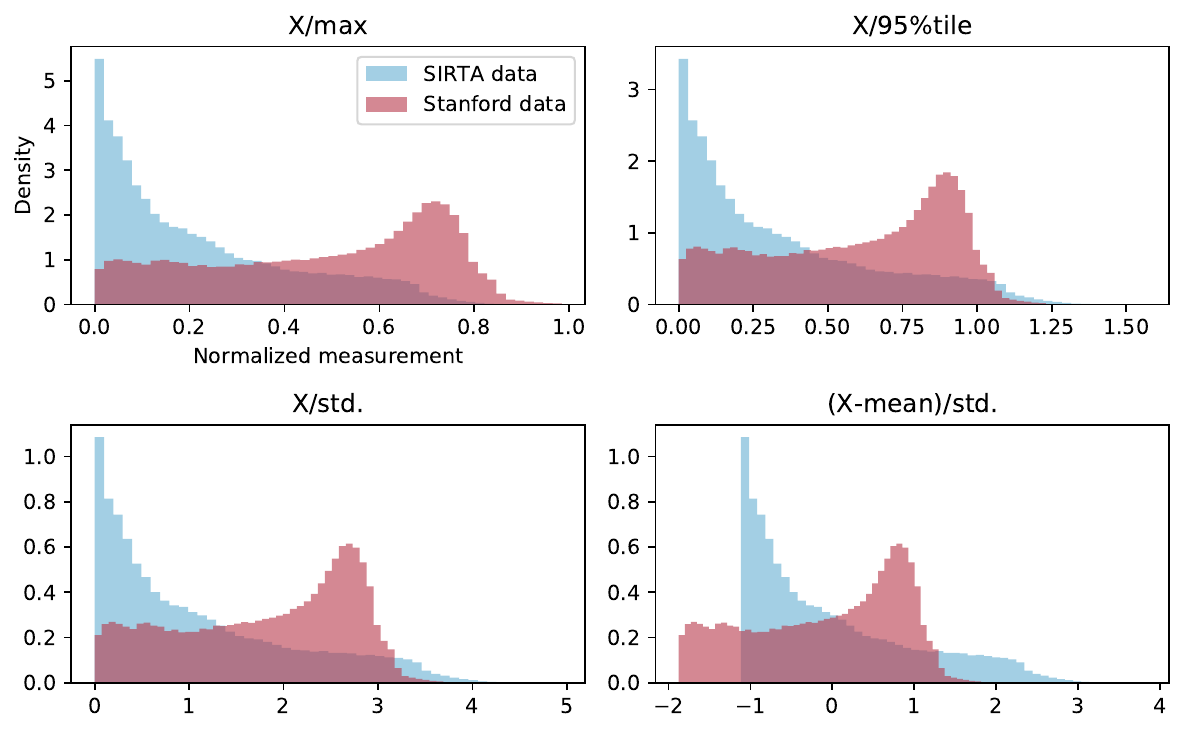}
\caption{The distribution of normalized measurements of the Stanford SIRTA datasets (Note: the distributions of $X/(max/100)$, $X/(95\%tile/100)$, $X/(std./100)$, $(X-mean)/(std./100)$ are not shown here because they have similar distributions as their counterparts and the only difference is the scale of 100x)}
\label{fig:normalized_pv_irradiance_distribution}
\end{figure}

To be clear, the normalization in this study is defined in the form of $(X-a)/b$, where $X$ is the data to be normalized, and $a$, $b$ are normalization factors. Two types of data are involved, namely, images and sensor measurements. For sky images, both teams normalized the pixel values of the images from $[0, 255]$ to $[0,1]$ by dividing the maximum pixel value 255, which is a common technique used for image data normalization. For PV output/irradiance measurements, including both input and output, different normalization methods are examined in this section and the details are summarized in Table \ref{tab:normalization_factors}. With ground truth prediction data being inaccessible when the model is deployed, the normalization factors are solely based on the input data, which is statistical similar. After the input and output of each dataset are normalized, we combine them, and the combined dataset is used for global model training. Figure \ref{fig:normalized_pv_irradiance_distribution} shows the data distribution of both datasets after normalization. In the testing phase, the same normalization factors are applied to the input data of the test set and the predictions are post-processed to the original scales. We evaluate the global models on the test sets of the two datasets individually and compare them with the corresponding local models.

\begin{table}[h!]
\begin{center}
\caption{Normalization methods in the form of $(X-a)/b$ for Stanford and SIRTA datasets}
\label{tab:normalization_factors}
\begin{small}
\begin{tabular}{>{\raggedright}p{0.4\linewidth}>{\raggedright}p{0.25\linewidth}>{\raggedright\arraybackslash}p{0.25\linewidth}}
\hline
Normalization method & Stanford & SIRTA \\
\hline\hline
\noalign{\vskip 1mm}
$X/max$ & $a=0,\hspace{1mm} b=30$ & $a=0,\hspace{1mm} b=1366$ \\
$X/95\%tile$ & $a=0,\hspace{1mm} b=24$ & $a=0,\hspace{1mm} b=853$ \\
$X/std.$ & $a=0,\hspace{1mm} b=8$ & $a=0,\hspace{1mm} b=270$ \\
$(X-mean)/std.$ & $a=15,\hspace{1mm} b=8$ & $a=301,\hspace{1mm} b=270$ \\
$X/(max/100)$ & $a=0,\hspace{1mm} b=0.3$ & $a=0,\hspace{1mm} b=13.36$ \\
$X/(95\%tile/100)$ & $a=0,\hspace{1mm} b=0.24$ & $a=0,\hspace{1mm} b=8.53$ \\
$X/(std./100)$ & $a=0,\hspace{1mm} b=0.08$ & $a=0,\hspace{1mm} b=2.7$ \\
$(X-mean)/(std./100)$ & $a=15,\hspace{1mm} b=0.08$ & $a=301,\hspace{1mm} b=2.7$ \\
\noalign{\vskip 1mm}
\hline
\end{tabular}
\end{small}
\end{center}

\footnotesize{Notes: $X$ represents the measurement values and $b$ is referred to as the normalization factor; $std.$ and $95\%tile$ represent standard deviation and 95 percentile of the data, respectively.}
\end{table}

\subsubsection{Architecture tuning}
\begin{figure}[h!]
\centering    
\makebox[\textwidth][c]{\includegraphics[width=1.2\textwidth]{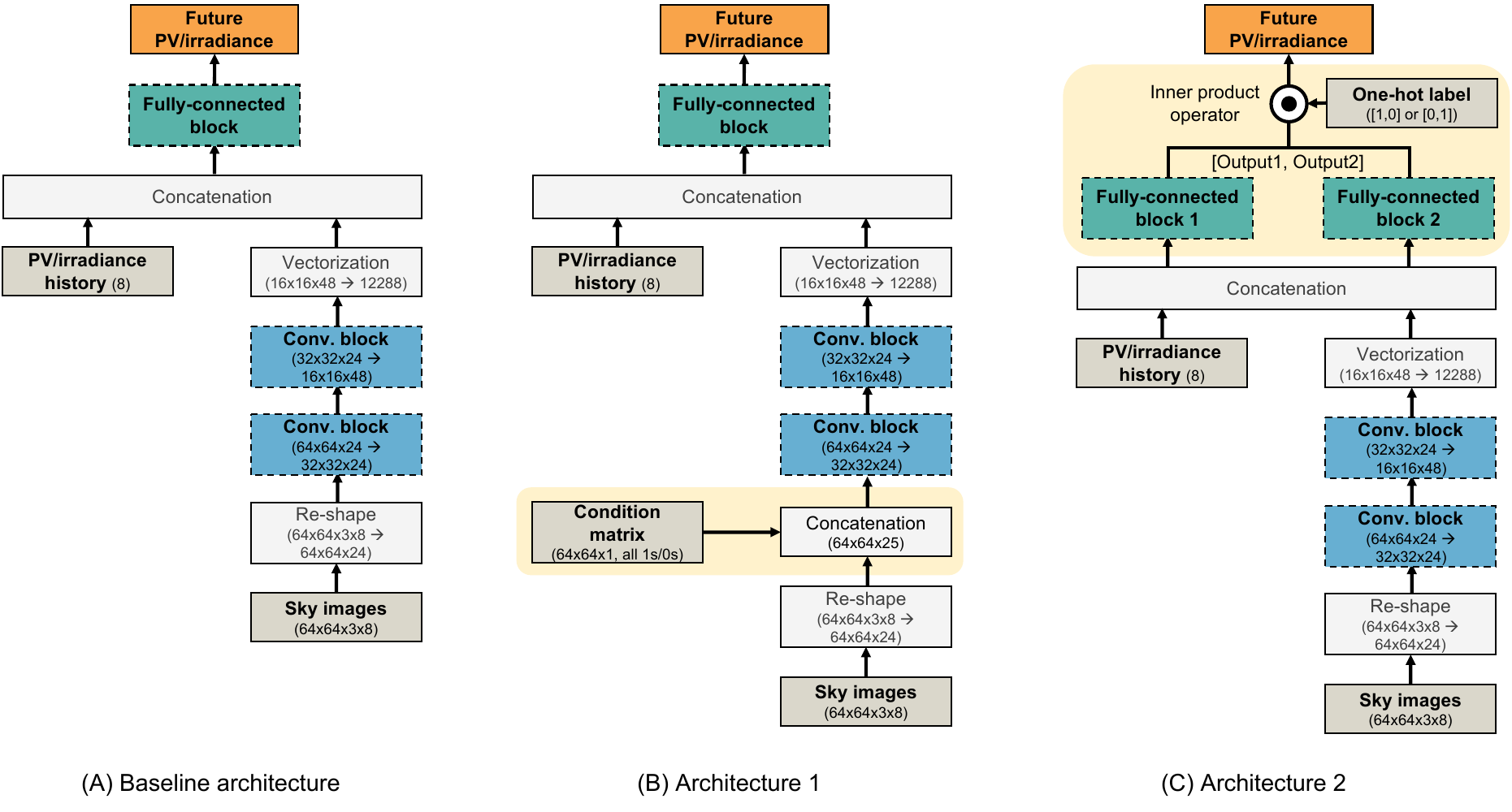}}
\caption{Illustration of different architecture set-ups using the SUNSET model as an example (the modifications from the baseline architecture are highlighted with yellow background)}
\label{fig:architecture_tuning}
\end{figure}

Two alternative architectures are investigated to deal with the two data input streams, i.e. the Stanford and SIRTA data, and are compared with the baseline architectures described in the Section \ref{subsec:deep learning models}. The main difference between the baseline and the two alternative architectures is that the location information is given explicitly to the alternative architectures to help the models distinguish between the input samples from different locations, while the baseline model attempts to learn the location-specific features implicitly from the data. Figure \ref{fig:architecture_tuning} shows an illustration of the two alternative architectures, which are named as Architecture 1 [Arch. 1, see Figure \ref{fig:architecture_tuning} (B)] and Architecture 2 [Arch. 2, see Figure \ref{fig:architecture_tuning} (C)]. The modifications from the baseline architecture are highlighted with yellow background. It should be noted that here we use the SUNSET model to explain the architecture modifications, while the same idea can be applied to the ConvLSTM model. It should also be noted that the models are trained with the normalized data based on the optimal normalized method identified for each team via experiments described in Section \ref{subsubsec:input and output normalization}.

Arch. 1 has minor modifications compared with the baseline. The only change is the introduction of a condition matrix, with the same resolution as the sky images ($64\times64$). All elements of the condition matrix are either 1 or 0, indicating the location of the data, for example, if the sky images come from the Stanford dataset, the condition matrix elements are all 0s, whereas if the sky images come from the SIRTA dataset, the condition matrix elements are all 1s.

Arch. 2 adds more complexity to the basline architecture. It has shared convolutional blocks and two separate fully-connected blocks, with the two fully-connected blocks consisting of the same components as the baseline architecture [see Figure \ref{fig:model_architecture} (A)]. The shared convolutional blocks act as a common feature extractor and learn features from both locations, while the two separate fully-connected blocks learn the correlation between the image features and the PV output/irradiance individually for each location, which is based on the fact that the sun angle trigonometry changes with locations. During the training process, each of the fully-connected blocks will generate a prediction and the two predictions are then stacked into a vector. To generate the final prediction, an inner product is computed between the prediction vector and the one-hot label vector, which indicates the location of the data, with [1,0] and [0,1] representing the Stanford and SIRTA data, respectively.

\subsection{Transfer learning}
\label{subsec:transfer_learning}
In this section, we examine if the knowledge, or specifically the feature representations from a pre-trained solar forecasting model can be leveraged, so that we do not have to train a model from scratch in a new location of interest. Here, we define a source dataset $\mathcal{D}_{S}$ and a corresponding source model $\mathcal{M}_{S}$, as well as a target dataset $\mathcal{D}_{T}$ and a target model $\mathcal{M}_{T}$. In most cases, $\mathcal{D}_{S}$ is a large dataset that contains massive samples, whereas $\mathcal{D}_{T}$ is limited in the number of samples. The goal of transfer learning, specifically, domain adaptation, is thus to learn $\mathcal{M}_{T}$ using $\mathcal{D}_{T}$ with the knowledge gained from learning $\mathcal{M}_{S}$ based on $\mathcal{D}_{S}$. It should be noted that the same architecture is used for $\mathcal{M}_{S}$ and $\mathcal{M}_{T}$ in this study, namely, either SUNSET by team Stanford or ConvLSTM by team Cambridge. 

To implement transfer learning, we first develop $\mathcal{M}_{S}$ based on $\mathcal{D}_{S}$. Two different strategies are then investigated for transferring the knowledge from $\mathcal{M}_{S}$ to $\mathcal{M}_{T}$: 
\begin{itemize}
    \item warm-starting strategy (WS): instead of initializing the weights of $\mathcal{M}_{T}$ randomly, it initializes the entire network with the weights of $\mathcal{M}_{S}$, and from that initial point, the training on $\mathcal{D}_{T}$ is started and new weights are learned; 
    \item freezing Conv. blocks strategy (FConv): the weights for all Conv. blocks from $\mathcal{M}_{S}$ are transferred and are frozen during the training process, while the weights of fully-connected blocks are initialized with the weights from $\mathcal{M}_{S}$ and learned with the target dataset $\mathcal{D}_{T}$. 
\end{itemize}  

\begin{figure}[h!]
\centering    
\makebox[\textwidth][c]{\includegraphics[width=1.0\textwidth]{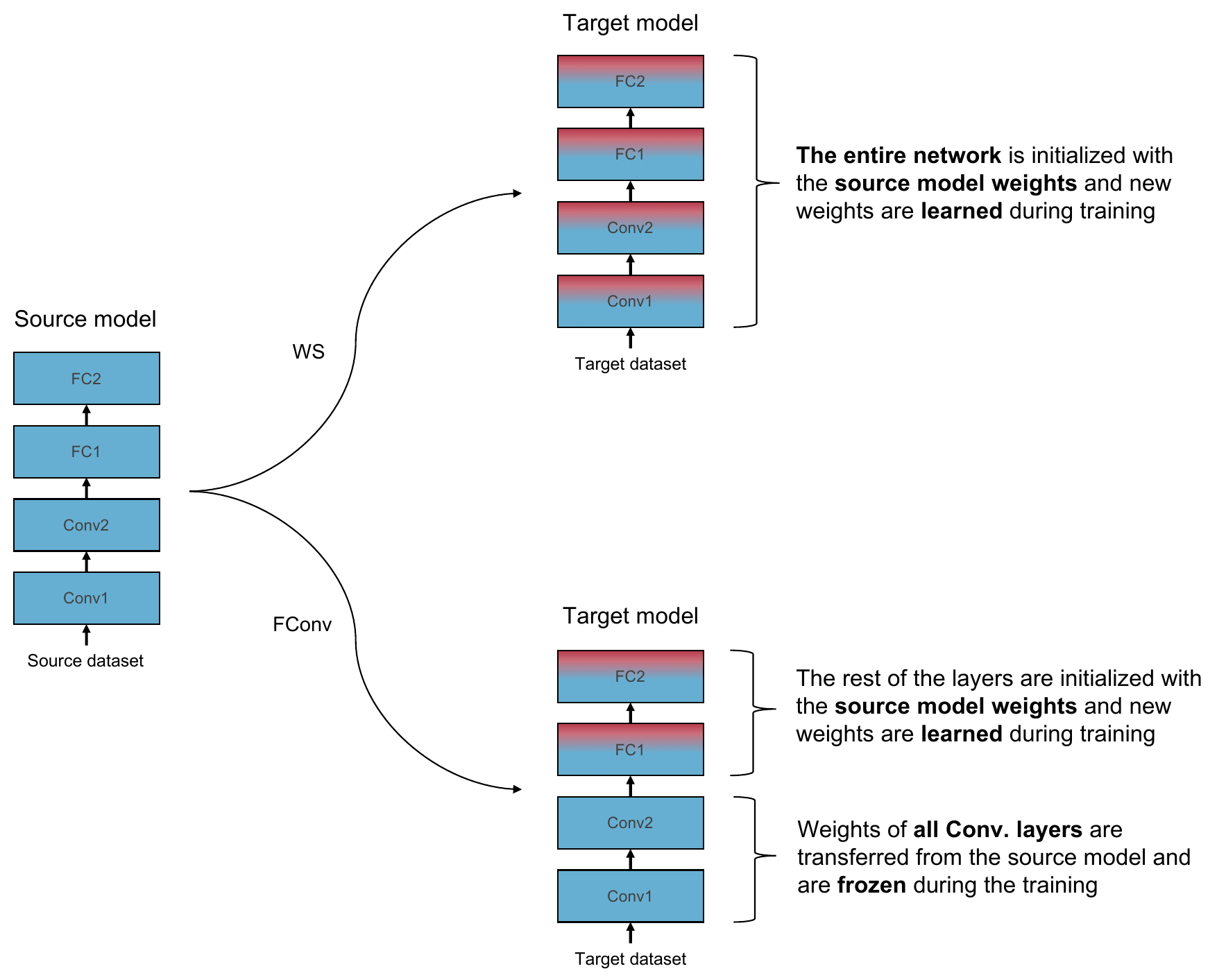}}
\caption{Illustration of different transfer learning strategies examined in this studies, taking a CNN with two convolutional layers (Conv1 and Conv2) and two fully connected layers (FC1 and FC2) as examples.}
\label{fig:illustration_transfer_learning}
\end{figure}

For FConv strategy, the feature extractors of $\mathcal{M}_{S}$  are reused by $\mathcal{M}_{T}$ and only the mapping functions associated with the fully connected layers are learned based on new data. An illustration of these two transfer learning strategies can be found in Figure \ref{fig:illustration_transfer_learning}.

Given that $\mathcal{D}_{S}$ generally has a wider data coverage than $\mathcal{D}_{T}$, we first use the DEWA dataset as $\mathcal{D}_{T}$, and experiment with the following combination of transfer learning: \{${\mathcal{D}_{S}}$: Stanford $\rightarrow {\mathcal{D}_{T}}$: DEWA\}, \{${\mathcal{D}_{S}}$: SIRTA $\rightarrow {\mathcal{D}_{T}}$: DEWA\} and \{${\mathcal{D}_{S}}$: Stanford+SIRTA $\rightarrow {\mathcal{D}_{T}}$: DEWA\}. We then examine the possibility of knowledge transfer between two sky image datasets with drastically different prediction targets and data distribution, namely, \{${\mathcal{D}_{S}}$: SIRTA $\rightarrow {\mathcal{D}_{T}}$: Stanford\} and \{${\mathcal{D}_{S}}$: Stanford $\rightarrow {\mathcal{D}_{T}}$: SIRTA\}. We also experiment with different amounts of training data for $\mathcal{D}_{T}$, including 1\%, 5\%, 10\%, 20\%, 50\%, 75\% and 100\%, roughly corresponding to 4, 16, 29, 55, 124, 184, 249 days of data for the DEWA development set, 6, 29, 59, 118, 272, 390, 502 days for the Stanford development set, and 9, 42, 79, 133, 329, 631, 933 days for the SIRTA development set. The data are sampled chronologically in the whole dataset to mimic the real situation of data collection. It should be noted that as team Cambridge does not have access to the DEWA dataset, they only conduct the second part of expriments and team Stanford (with access to the DEWA dataset) conducts all the experiments mentioned above. In this study, both the source models $\mathcal{M}_{S}$ and the target models $\mathcal{M}_{T}$ are trained using the normalized data rather than the original data, again based on the optimal normalization method identified by each team. All transfer learning models are evaluated and compared with the local model baseline trained from scratch using $\mathcal{D}_{T}$ in two aspects: prediction accuracy measured by the RMSE, and training cost measured by the TE as described in Section \ref{subsec:eval_metrics}. 

\section{Results and discussion}
\label{sec:results and discussion}

\subsection{Comparison of local and global models for solar forecasting}
\label{subsec:compare_local_and_global_model_results}
We first compare the performance of local and global solar forecasting models. Besides serving as local baselines, the local models are also applied offsite without re-training by predicting the relative values of irradiance/PV output and post-processing to the original scales for performance evaluation. A global model is trained using the combination of normalized Stanford and SIRTA dataset. The normalization methods used for both local models and global models are $X/(std./100)$ for team Stanford and $X/95\%tile$ for team Cambridge, which are the optimal normalization methods identified in the experiments presented in Section \ref{subsec:global_model_results}.

Table \ref{tab:local_global_model_performance} shows the overall test set performance of local and global models trained with baseline SUNSET and ConvLSTM architectures. For both architectures, the local models achieve the best performance when applied locally, and the prediction errors significantly increase when applied offsite (without re-training). Although there are still large prediction errors when applying local models to other locations with similar measurement data distributions, e.g., applying the local model developed using Stanford data (PV output) to the DEWA test set (GHI) (Test RMSE 155.82 W/m$^2$), it tends to have better performance than applying local models to locations with quite different data distribution even though they share the same target variable, e.g., applying the local model developed using SIRTA data (GHI) to DEWA test set (GHI) (Test RMSE 219.79 W/m$^2$). 

\begin{figure}[h!]
\centering    
\makebox[\textwidth][c]{\includegraphics[width=1.0\textwidth]{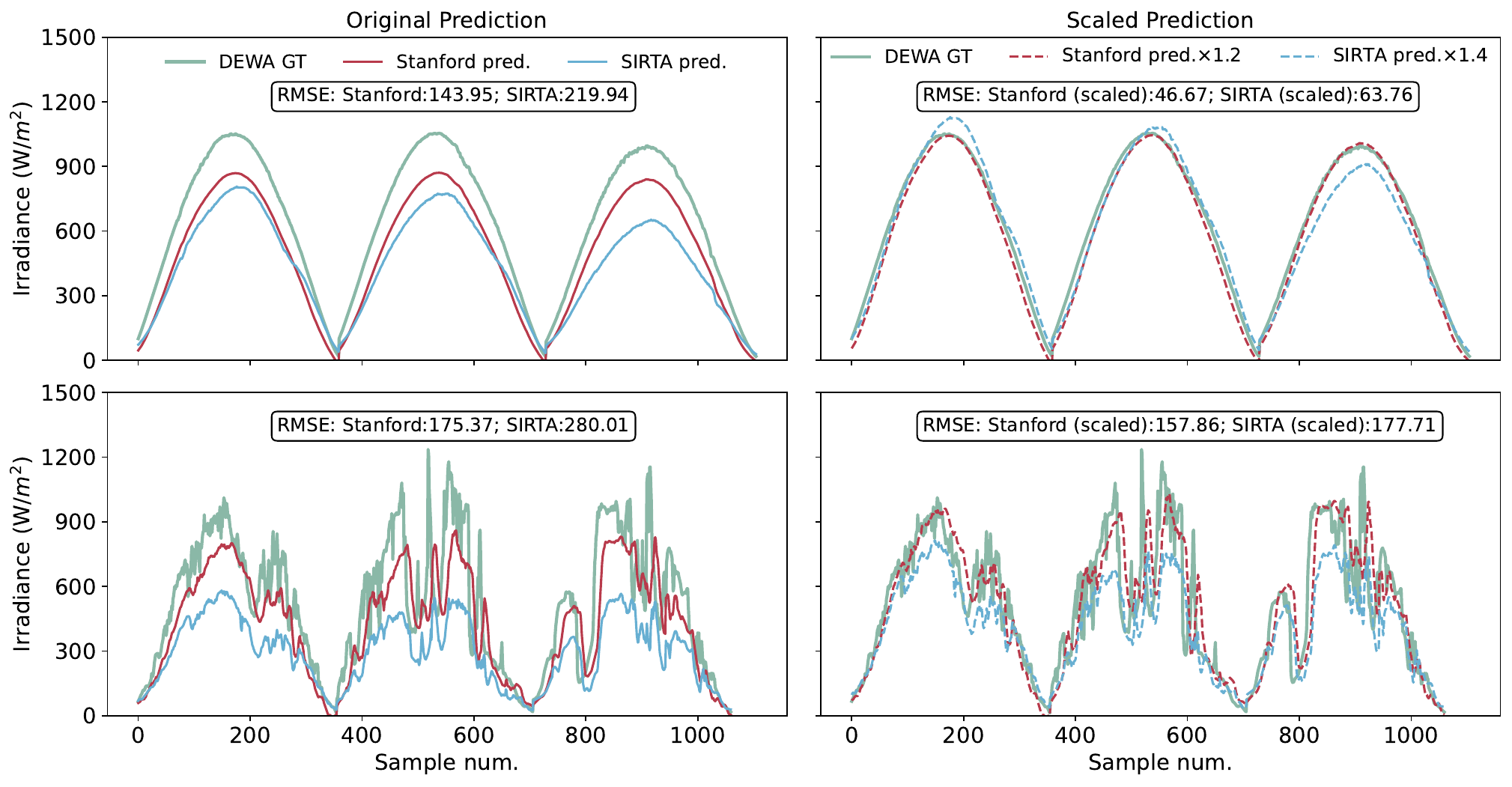}}
\caption{Visualization of Stanford and SIRTA local model predictions based on the SUNSET architecture applied to 6 example days (3 sunny days shown in the first row and 3 cloudy days shown in the second row) in the DEWA test set without re-training. The first column shows the original predictions and the second column shows the scaled predictions from simply multiplying the original prediction by a factor. The same factor is applied for both sunny days and cloudy days predictions. (GT: ground truth; pred.: prediction)}
\label{fig:local_model_result_examples}
\end{figure}

The prediction errors of applying learned local models offsite can be largely attributed to two parts of the model architectures, the feature extraction part associated with the Conv. blocks and the regression part associated the fully-connected blocks. We find that the errors mainly come from the regression part, namely, mapping extracted features with the prediction target values, whereas the feature representations learned based on each individual dataset can to some extent be shared. This is illustrated in Figure \ref{fig:local_model_result_examples} which shows the predictions of local models developed based on Stanford and SIRTA datasets individually using the SUNSET architecture applied to 6 example days (3 sunny and 3 cloudy days) in the DEWA test set without re-training. The first column of the figure shows the original predictions, and the second column shows the scaled predictions obtained by multiplying the original predictions by a factor of 1.2 for the Stanford local model and 1.4 for the SIRTA local model for both example sunny and cloudy days, without any other treatment. The original prediction curves, including both Stanford and SIRTA local model predictions, tend to have similar shapes as the ground truth curves, regardless of sunny days and cloudy days, which suggests that feature representations learnt by the models are common for all the locations and can be shared. The errors are mostly caused by the scale or magnitude of the predictions. Once we manually scale the prediction by a factor, we can observe significant improvement in the prediction, although it is not perfect. This finding is further illustrated in Section \ref{subsec:transfer_learning_results} by freezing the weights of the Conv. blocks and only training the fully connected blocks during transfer learning.

\begin{table}[h!]
\begin{adjustbox}{center,angle=0}
\begin{threeparttable}
\caption{Overall test set performance of local and global models (the best prediction performance on each local test set is highlighted in bold font)}
\begin{small}
\begin{tabular}{p{0.13\linewidth}|p{0.2\linewidth}|p{0.10\linewidth}|p{0.1\linewidth}|p{0.14\linewidth}|p{0.1\linewidth}|p{0.1\linewidth}|p{0.1\linewidth}}
\hline
\multirow{2}{*}{Model} & \multirow{2}{*}{Trained on} & \multirow{2}{2cm}{Training epochs} & \multirow{2}{1.8cm}{Training samples}  & \multirow{2}{1.8cm}{Training effort} & \multicolumn{3}{c}{Test RMSE on} \\ \cline{6-8}
      &   &  &  &  & Stanford & SIRTA & DEWA \\
\hline\hline
\noalign{\vskip 1mm}
\multirow{4}{*}{SUNSET} & Stanford & 21.2$\pm$6.1 & 125,876 & 2.67$\pm$0.77M & 2.68 & 136.96 & 155.82 \\
       & SIRTA & 9.6$\pm$1.8 & 438,172 & 4.21$\pm$0.81M & 6.53 & 96.42 & 219.79 \\
       & DEWA &  11.8$\pm$2.2 & 85,278 & 1.01$\pm$0.19M & 4.65 & 160.28 & \textbf{89.49} \\
       & Stanford+SIRTA & 28.6$\pm$5.6 & 564,048 & 16.13$\pm$3.15M & \textbf{2.61} & \textbf{96.34} & 129.21 \\
\noalign{\vskip 1mm}
\hline
\noalign{\vskip 1mm}

\multirow{3}{*}{ConvLSTM} & Stanford & 11.7$\pm$2.7 & 125,876 & 1.47$\pm$0.34M & 2.62 & 113.05 & N/A \\
       & SIRTA & 5.6$\pm$2.6 & 438,172 & 2.49$\pm$1.13M & 4.39 & 96.46 & N/A \\
       & Stanford+SIRTA & 6.0$\pm$1.7 & 564,048 & 3.36$\pm$0.99M & \textbf{2.62} & \textbf{95.53} & N/A \\
\noalign{\vskip 1mm}
\hline     
\end{tabular}
\label{tab:local_global_model_performance}
\begin{tablenotes}
\item Notes: (1) The training effort reported in this table is defined as the number of samples seen by the model until convergence, i.e. the number of training epochs $\times$ the number of training samples, representing the average performance (mean$\pm$std) for training one sub-model from ten-fold cross-validation. M stands for million samples. (2) The training epochs are obtained when the model stops training, namely, the validation loss is not observed to decrease for five consecutive epochs (3) The test RMSE is calculated based on the ensemble mean prediction of the ten sub-models from ten-fold cross-validation.
\item
\end{tablenotes}
\end{small}
\end{threeparttable}
\end{adjustbox}
\end{table}

In contrast, the global model trained with a combined dataset from two locations can perform well on individual test sets from both locations. The performance is close to, or even better than, the corresponding local models (e.g., Stanford+SIRTA global model \textit{versus} Stanford local model \textit{versus} SIRTA local model), which suggests that the global models can learn features from both locations simultaneously and the learned features can be correlated with the prediction targets in a relatively separate fashion without compromising the performance for each location. This property is further examined in Section \ref{subsec:global_model_results} by comparing the baseline architectures with the modified architectures that explicitly disentangle the location information. Moreover, including diversified samples from both locations could improve the model generalization ability. 

Although applying Stanford+SIRTA global model to the DEWA test set gives much worse performance than the DEWA local model, it is significantly better than the Stanford and SIRTA local models applied to the DEWA test set. We could reasonably expect training a global model on a combination of Stanford, SIRTA and DEWA sets could give promising results on each of the three datasets.

In terms of the prediction accuracy, training a global model with a combined dataset and applying it to locations of interest is superior, especially if the local datasets have limited sample size. However, the training cost of global models remains a challenge. Depending on specific model architectures, more efforts might be required in the training process, due to increases in both training set size and training epochs, for the model to converge. A further evaluation on the training cost is presented in Section \ref{subsec:overall_evaluation} for comparison of local, global models, and models trained with transfer learning.

\subsection{A further look into global solar forecasting models}
\label{subsec:global_model_results}
In view of the superior prediction accuracy of the global models over the local models, in this section, we further present experiments on training global models given the dataset heterogeneity, especially the different scales and distributions of prediction targets. Also, we experiment with different alternative model architectures to accommodate the multi-location input data, and compare these with the baseline model architectures.

\begin{figure}[h!]
\centering    
\makebox[\textwidth][c]{\includegraphics[width=1.0\textwidth]{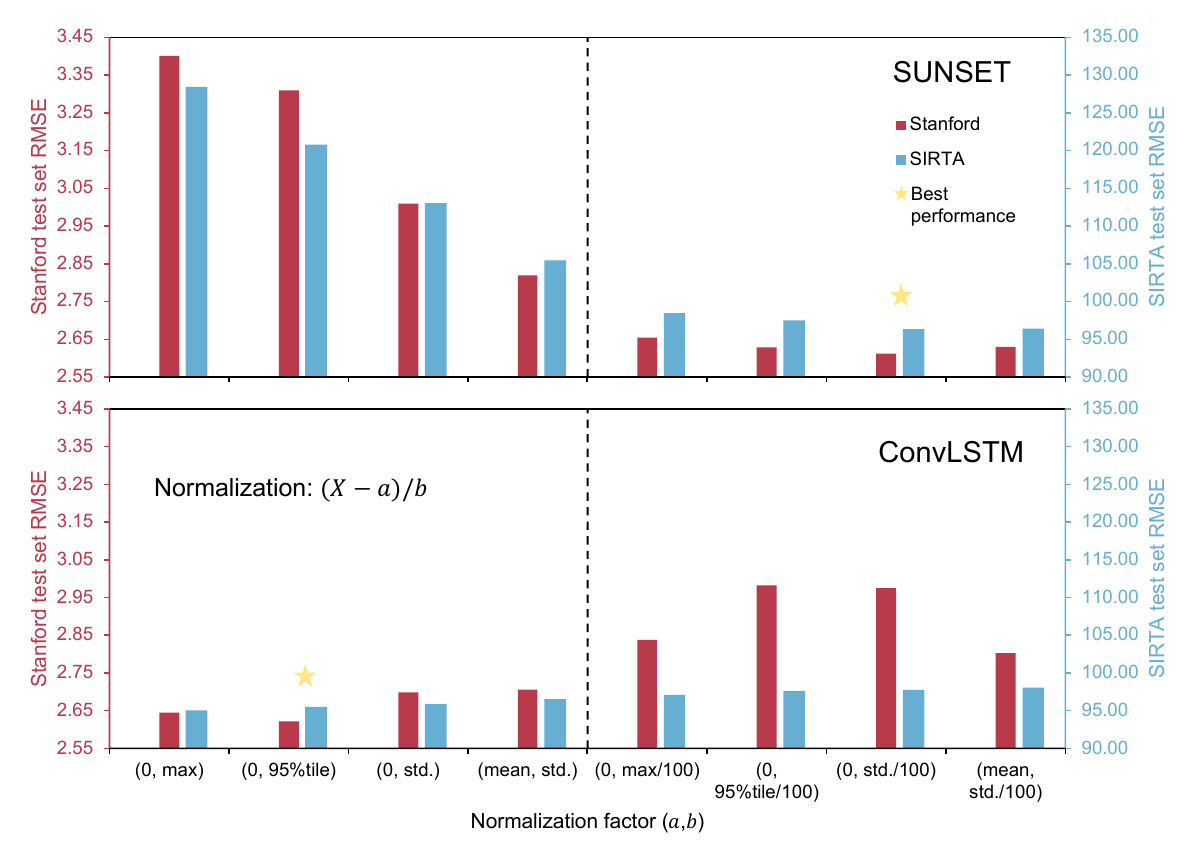}}
\caption{Comparison of different normalization methods for training global the model for two different model architectures}
\label{fig:differet_normalization_methods_results}
\end{figure}

Figure \ref{fig:differet_normalization_methods_results} shows the comparison of different normalization methods for training the global model for SUNSET and ConvLSTM architectures, respectively. The global model is trained jointly using the combination of Stanford and SIRTA datasets and evaluated individually on the two local datasets. It can be observed that the SUNSET model is more sensitive to the scale of the normalization factors compared with the ConvLSTM model, dividing the normalization factor by 100 can significantly improve the model performance on both local datasets. In contrast, the ConvLSTM model behaves such that a small normalization factor generally leads to worse performance. The optimal normalization factors for SUNSET and ConvLSTM are identified as $(0,std./100)$ and $(0,95\%tile)$, respectively. It should be noted that normalization factors derived from robust statistics, e.g. $std$ or $95\%tile$, are less affected by outliers compared to, say, the $max$ value. Preliminary experiments suggest that the different responses to the normalization factors are due to the different architectures of the two models, especially the batch normalization layers (SUNSET uses batch normalization layers but ConvLSTM does not use), though the underlying reasons need to be more thoroughly understood via future studies. We also suggest that researchers should take care in deciding which normalization methods they use for their models.

The results of including dataset location as an auxiliary data input (see Figure \ref{fig:architecture_tuning} for a comparison of different global model architectures) are presented in Table \ref{tab:differet_global_model_architecture_results}. No significant performance improvement is observed for the two alternative architectures for both SUNSET and ConvLSTM models compared with the baseline architectures, which suggests that the baseline model can learn by itself to distinguish location using the subtle features with the prediction target. Thus, the modifications in the architecture to manually inject the location information or disentangle the prediction for each location are not necessary.

\begin{table}[h!]
\begin{center}
\caption{Performance of different global model architectures on local test sets (the best performance is highlighted in bold font)}
\begin{small}
\begin{tabular}{p{0.14\linewidth}|p{0.16\linewidth}|p{0.15\linewidth}|p{0.2\linewidth}|p{0.10\linewidth}|p{0.08\linewidth}} \hline
\multirow{2}{*}{Model} & \multirow{2}{1.2cm}{Normalization method} & \multirow{2}{1.2cm}{Architecture choice} & \multirow{2}{*}{Trained on} & \multicolumn{2}{c}{Test RMSE on} \\ \cline{5-6}
 & & & & Stanford & SIRTA \\
\hline\hline
\noalign{\vskip 1mm}
\multirow{3}{*}{SUNSET} & \multirow{3}{*}{$X/(std./100)$} & Baseline & Stanford+SIRTA & 2.612 & 96.335  \\
 & & Arch. 1 & Stanford+SIRTA & \bf{2.606} & \bf{96.075}  \\
 & & Arch. 2 & Stanford+SIRTA  & 2.610 & 96.585  \\
\noalign{\vskip 1mm}
\hline
\noalign{\vskip 1mm}
\multirow{3}{*}{ConvLSTM} & \multirow{3}{*}{$X/95\%tile$} & Baseline & Stanford+SIRTA & \bf{2.622} & \bf{95.529} \\
 & & Arch. 1 & Stanford+SIRTA & 2.765 & 96.850 \\
 & & Arch. 2 & Stanford+SIRTA  & 2.663 & 96.264 \\
\noalign{\vskip 1mm}
\hline     
\end{tabular}
\end{small}
\end{center}
\label{tab:differet_global_model_architecture_results}
\end{table}

\subsection{Knowledge transfer for solar forecasting}
\label{subsec:transfer_learning_results}

\begin{figure}[h!]
\centering    
\makebox[\textwidth][c]{\includegraphics[width=1.0\textwidth]{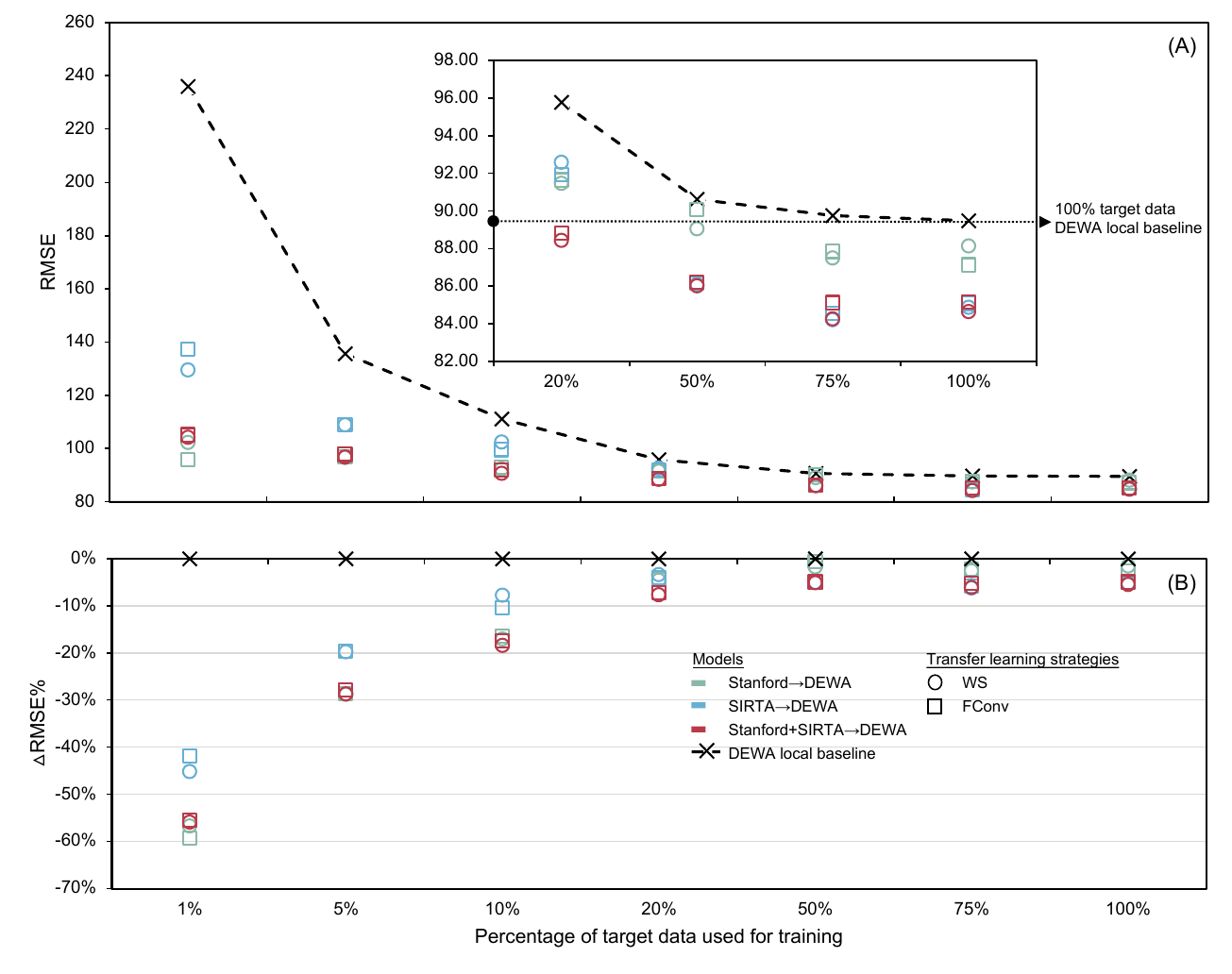}}
\caption{Comparison of different transfer learning strategies and different amounts of data available for the target dataset. The RMSEs are evaluated based on the DEWA test set. Subplot (A) shows the RMSE for models learned using different transfer learning strategies at each level of target data; The insert plot in (A) shows the zoom-in of results from 20\% to 100\% target data and the dotted line represents the RMSE of the DEWA local baseline model trained using 100\% data; The baseline performance is represented by the dashed line in each subplot; Subplot (B) shows the relative performance of the models learned using different transfer learning strategies compared with the DEWA local model baseline for each level of target data. For each target data level, local models correspond to $\Delta$RMSE\% = 0\%, and negative $\Delta$RMSE\% indicates that the models learned with different transfer learning strategies outperform the corresponding DEWA local model baseline.}
\label{fig:transfer_learning_strate_results}
\end{figure}

\begin{figure}[h!]
\centering    
\makebox[\textwidth][c]{\includegraphics[width=1.3\textwidth]{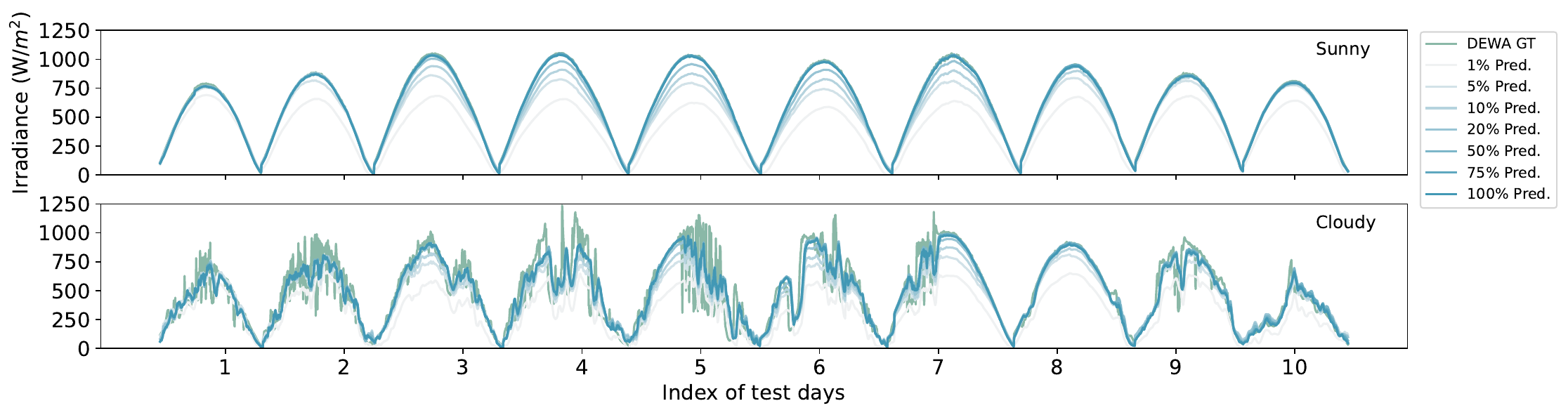}}
\caption{Visualization of predictions (Pred.) from the DEWA local baseline model trained with different amounts of data using the SUNSET architecture versus ground truth (GT) on the DEWA test set. The green curve represents the ground truth. Different shades of blue represent the local models trained with different amounts of data (light blue stands for a small dataset for training, dark blue stands for a large dataset for training). Note that for sunny days, the predictions from model trained with 50\% data and above almost overlap with the ground truth. Huge gaps can be observed between model predictions and ground truth if there is less than 20\% data for training.}
\label{fig:dewa_basline_diff_data}
\end{figure}

Lastly, we explore the potential of transferring knowledge from a pre-trained solar forecasting model developed from a source dataset to a new model based on a target dataset. 

We first present the experiments using the DEWA dataset as the target dataset, and the knowledge is transferred from the models learned from source datasets Stanford, SIRTA and Stanford+SIRTA, respectively. Figure \ref{fig:transfer_learning_strate_results} shows the results of transfer learning with different strategies (WS and FConv) and different amounts of target data as described in Section \ref{subsec:transfer_learning}. All models for this experiment are trained based on the SUNSET architecture and the performance is evaluated based on the DEWA test set.

In general, models trained using transfer learning strategies, no matter WS or FConv, achieve better prediction performance than the DEWA local baseline models for every level of target data used. The smaller the local dataset is, the more benefit transfer learning can bring. When target data is less than 20\% of the whole DEWA dataset (55-day-equivalent amount of data), the transfer learning models can significantly outperform the local baseline by approximately 10\% and up to 60\%. An extreme case is when there is only 1\% of total data (4-day-equivalent amount of data). The transfer learning models can outperform the baseline by nearly 60\%, which indicates that the knowledge gained from training models on other large datasets can be transferred and reused in the new task for a jump-start. From 20\% target data and beyond, the benefits that transfer learning can bring are gradually diminished. Although it still outperforms the baseline, the improvement is within 10\%.

In terms of different transfer learning strategies, namely, WS and FConv, the difference in prediction performance is generally less than 1\%. Although in some specific situations (e.g., 1\% target data), a 6\% difference can be observed, generally, there is no significant advantage for using one strategy over another. To this end, freezing all the Conv. blocks during the model training is totally acceptable, as the learned feature representations from the source models can largely be reused in the target model without compromising the performance. Even when the data distribution is drastically different, e.g., the SIRTA dataset \textit{versus} the DEWA dataset, some common features learned by the source model can be shared.

Moreover, by comparing different source models for transferring the knowledge, it can be observed that the larger and the more diversified the dataset used for building the source model, the more benefit it can bring in the transfer learning process. Using the Stanford+SIRTA dataset to build the source model can take advantage of the two individual datasets. When target data is less than 20\%, the Stanford$\rightarrow$DEWA model shows better performance than the SIRTA$\rightarrow$DEWA model as a similar data distribution is shared by the Stanford and DEWA datasets and thus the learned features can be easily transferred, while when target data is greater than 20\%, the opposite can be observed. When target data is greater than 20\%, it is enough for the model to learn a solid sun angle equation (see Figure \ref{fig:dewa_basline_diff_data}, when training data is greater than 20\%, the sunny day predictions almost align with the ground truth), and the overall performance of the model on the test set is mainly determined by the prediction accuracy on cloudy days. Therefore, the features learned from the SIRTA dataset which is dominant in cloudy data can contribute more. Another result is the huge data reduction potential by using transfer learning. It can be noticed that with 20\% target data the Stanford+SIRTA$\rightarrow$DEWA transfer learning model can even achieve slightly better performance than the case of 100\% target data using the DEWA local baseline model.

\begin{figure}[h!]
\centering    
\makebox[\textwidth][c]{\includegraphics[width=1.1\textwidth]{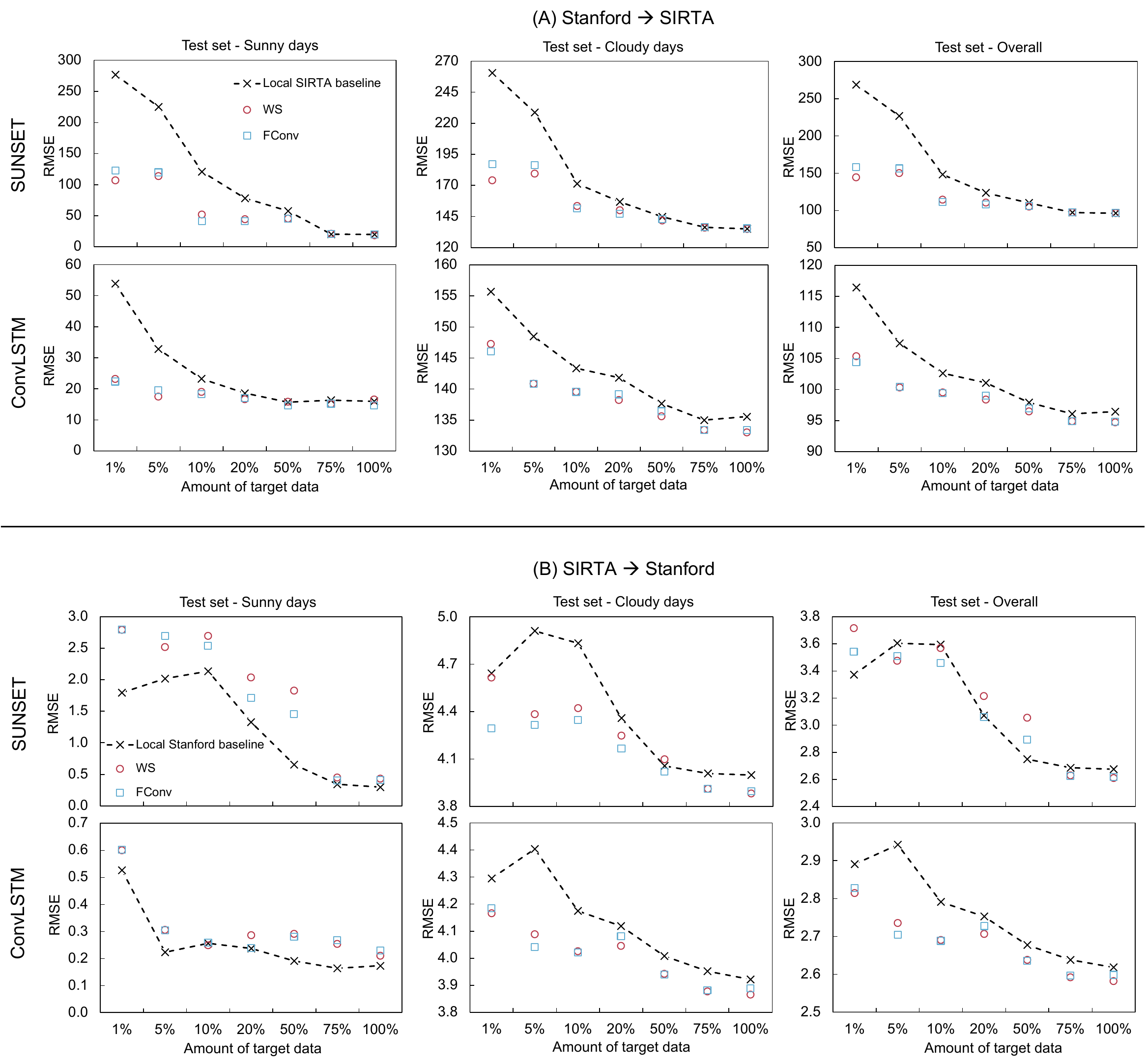}}
\caption{Transfer learning between Stanford dataset and SIRTA dataset for SUNSET and ConvLSTM models with different amounts of target data and different transfer learning strategies. Subplot (A) shows the models pre-trained on the Stanford dataset and transferred to the SIRTA dataset using different strategies; Subplot (B) shows the models pre-trained on the SIRTA dataset and transferred to the Stanford dataset using different strategies.}
\label{fig:transfer_learning_stanford_sirta}
\end{figure}

Transfer learning from SIRTA to DEWA shows promising results as described above, although the two datasets have very different data distribution. Here, we further examine two even harder cases, namely, transfer learning from Stanford to SIRTA and from SIRTA to Stanford, with differences both in prediction target (PV output versus GHI) and data distribution. The Stanford dataset contains only approximately a quarter of the data volume of the SIRTA dataset, and has relatively more balanced sunny and cloudy sample distribution than the SIRTA dataset, which is dominated by cloudy data. Figure \ref{fig:transfer_learning_stanford_sirta} shows transfer learning between the Stanford and the SIRTA datasets for the SUNSET and ConvLSTM models with different amounts of target data and different transfer learning strategies.

In general, SUNSET and ConvLSTM show similar trends in both transfer learning cases. From Stanford$\rightarrow$SIRTA, the transfer learning models outperform the local models on both sunny and cloudy days, while the benefits gradually saturate when more training data is available, which is similar to the previous experiments on transfer learning to the DEWA dataset. It also suggests that the knowledge learned from PV power prediction can be transferred to irradiance prediction as these two variables are correlated with each other. However, a caveat should be added here. PV power output is affected by the panel temperature (i.e. high temperature can hinder the power generation of PV), which could be learned by the deep learning model from the data, while irradiance measurements are less affected by temperature changes. Therefore, the temperature parameter needs to be taken into account for transfer learning from PV to irradiance. The average temperature in Stanford is below 30$^{\circ}$C in the summer, hence its impact is not observed in this study. From SIRTA$\rightarrow$Stanford, different behaviors are observed on sunny and cloudy days. The local models outperform the transfer learning models on sunny days, while the transfer learning models do better on cloudy days. This makes sense as the knowledge transferred from the source model (SIRTA in this case) is mostly trained on cloudy data. The WS or FConv transfer learning strategy could put the model at a sub-optimal starting point and hinder the learning if there is not enough data, while things can become better as more training data is available. In comparison, Stanford dataset has a relative more balanced distribution of sunny and cloudy samples. It can be observed that transfer learning from Stanford to SIRTA outperforms the SIRTA local baseline. It thus suggests that for transfer learning, the balance of dataset samples should be paid attention to besides purely the size of the dataset.

\subsection{Overall evaluation of different training strategies}
\label{subsec:overall_evaluation}
Table \ref{tab:compare_local_with_transfer_learning_model} summarizes the performance of local, global and transfer learning models on each one of the three datasets for both SUNSET and ConvLSTM architectures. The performance is evaluated based on both the prediction accuracy (measured by target test dataset RMSE) as well as the training cost (measured by training effort as defined in Section \ref{subsec:eval_metrics}). For the DEWA dataset as the target set, the results show that by using transfer learning, either WS or FConv, with 80\% less training data and training effort (TE), we can achieve 1\% improvement over the local baseline model trained with the whole dataset. If the whole target training data is available, models trained with transfer learning can outperform the local model by 5\% with 14\% less training effort. For the Stanford dataset and the SIRTA dataset as the target set, both SUNSET and ConvLSTM architectures are used in the experiments. The two architectures show different behaviors in model convergence, which leads to differences in TE, while the general trend is aligned. For both architectures, with the whole dataset available, transfer learning models can achieve slightly better prediction accuracy than the local model and global model (improvement within 3\%). Depending on the model and the direction of the transfer between SIRTA and Stanford datasets, transfer learning strategies impact the TE differently. For the SUNSET model, the TE of transfer learning relative to the baseline local model ranges from $-6$ to $+10$\%, while for the ConvLSTM model it decreases from $-4$ to $-87$\%. In particular, the TE is down by a factor of 8 to 9 when ConvLSTM is used to transfer knowledge from SIRTA to Stanford. Although the global model is more generalized and can be applied to individual locations without compromising its prediction ability (as presented in Table \ref{tab:local_global_model_performance}), it generally needs more training effort than the local model and the transfer learning model due to the increasing dataset size and possibly more training epochs to converge.

Overall, with the superior performance in terms of both prediction accuracy and computational cost in model training, transfer learning appears to be the best of the three training strategies. It is especially suitable for the situation where there is limited data. As more and more open-source datasets are available to the community, we highly recommend joint efforts in building pre-trained solar forecasting models on a centralized large-scale dataset with massive and diversified samples (just like in computer vision, different pre-trained models are built using ImageNet \cite{dengImageNetLargescaleHierarchical2009}), which can potentially accelerate the research and development of solar forecasting methods. Local retraining starting from this baseline will allow superior local models.

\begin{table}
\begin{adjustbox}{center}
\begin{threeparttable}
\caption{Summary of the performance of local, global and transfer learning (TL) models for SUNSET and ConvLSTM architecture. The performance is evaluated for each of the dataset based on both the prediction accuracy (measured by target test set RMSE) and the computational cost of training (measured by training effort).}
\begin{small}
\begin{tabular}{>{\raggedright}p{0.14\linewidth}|>{\raggedright}p{0.21\linewidth}|>{\raggedright}p{0.21\linewidth}|>{\raggedright}p{0.08\linewidth}|>{\raggedright}p{0.1\linewidth}|>{\raggedright}p{0.13\linewidth}|>{\raggedright}p{0.10\linewidth}|>{\raggedright}p{0.10\linewidth}|>{\raggedright\arraybackslash}p{0.14\linewidth}}
\hline
Model architecture & Target training dataset (\% data) & Source training dataset  & Model type & TL strategy & Target test set RMSE & Training samples & Training epochs & Training effort \\ 
\hline\hline
\noalign{\vskip 1mm}
\multirow{6}{*}{SUNSET} & \multirow{3}{2cm}{DEWA (20\%)} & N/A & L & N/A & 95.76 & 17,056  & 23.4$\pm$3.4 & 0.40$\pm$0.06M \\ \cline{3-9}
       &  & \multirow{2}{3cm}{Stanford +SIRTA} & \multirow{2}{1cm}{TL} & WS & 88.44 & 17,056 & 12.7$\pm$2.9  & 0.22$\pm$0.05M \\
       &  &  &  & FConv & 88.82 & 17,056 & 12.8$\pm$5.6 & 0.22$\pm$0.10M \\ \cline{2-9}
 & \multirow{3}{3cm}{DEWA (100\%)} & N/A & L & N/A & 89.49 & 85,278 & 11.8$\pm$2.2 & 1.01$\pm$0.19M \\ \cline{3-9}
&  & \multirow{2}{3cm}{Stanford +SIRTA} & \multirow{2}{1cm}{TL} & WS & 84.64  & 85,278 & 10.9$\pm$2.4 & 0.93$\pm$0.21M  \\
&  &  &  & FConv & 85.13  & 85,278 & 10.1$\pm$4.6 & 0.86$\pm$0.39M  \\ 
\noalign{\vskip 1mm}
        \hline
\noalign{\vskip 1mm}
\multirow{8}{*}{SUNSET}       & \multirow{4}{3cm}{Stanford (100\%)} & N/A & L & N/A & 2.68  & 125,876 & 21.2$\pm$6.1 & 2.67$\pm$0.77M  \\ \cline{3-9}
       &  & Stanford +SIRTA & G & N/A & 2.61  & 564,048 & 28.6$\pm$5.6 & 16.13$\pm$3.15M  \\ \cline{3-9}
       &  & \multirow{2}{1cm}{SIRTA} & \multirow{2}{1cm}{TL} & WS &  2.61 & 125,876 & 23.2$\pm$7.7 & 2.92$\pm$0.97M  \\
       &  &  &  & FConv & 2.62  & 125,876 & 22.7$\pm$6.9 & 2.86$\pm$0.87M  \\ \cline{2-9}
       & \multirow{4}{3cm}{SIRTA (100\%)} & N/A & L & N/A & 96.42  & 438,172 & 9.6$\pm$1.8 &  4.21$\pm$0.81M \\ \cline{3-9}
       &  & Stanford +SIRTA & G & N/A & 96.34  & 564,048 & 28.6$\pm$5.6 & 16.13$\pm$3.15M  \\ \cline{3-9}
       &  & \multirow{2}{1cm}{Stanford} & \multirow{2}{1cm}{TL} & WS & 96.18 & 438,172  & 9.0$\pm$2.0 & 3.94$\pm$0.88M   \\
       &  &  &  & FConv & 96.73  & 438,172 & 10.6$\pm$2.6 & 4.64$\pm$1.15M  \\ 
\noalign{\vskip 1mm}
\hline
\noalign{\vskip 1mm}
\multirow{8}{*}{ConvLSTM} & \multirow{4}{3cm}{Stanford (100\%)} & N/A & L & N/A & 2.62 & 125,876 & 11.7$\pm$2.7 & 1.47$\pm$0.34M \\ \cline{3-9}
       &  & Stanford +SIRTA & G & N/A & 2.62 & 564,048 & 6.0$\pm$1.7 & 3.36$\pm$0.99M \\ \cline{3-9}
       &  & \multirow{2}{1cm}{SIRTA} & \multirow{2}{1cm}{TL} & WS & 2.58 & 125,876 & 1.5$\pm$0.9 & 0.19$\pm$0.11M \\
       &  &  &  & FConv & 2.60 & 125,876 & 1.7$\pm$1.4 & 0.22$\pm$0.17M \\ \cline{2-9}
       & \multirow{4}{3cm}{SIRTA (100\%)} & N/A & L & N/A & 96.46 & 438,172 & 5.6$\pm$2.6 & 2.49$\pm$1.13M \\ \cline{3-9}
       &  & Stanford +SIRTA & G & N/A & 95.53 & 564,048 & 6.0$\pm$1.7 & 3.36$\pm$0.99M \\ \cline{3-9}
       &  & \multirow{2}{1cm}{Stanford} & \multirow{2}{1cm}{TL} & WS & 94.73 & 438,172 & 5.4$\pm$1.6 & 2.38$\pm$0.70M \\
       &  &  &  & FConv & 94.83 & 438,172 & 4.5$\pm$1.3 & 1.98$\pm$0.56M \\ 

\noalign{\vskip 1mm}
\hline     
\end{tabular}
\label{tab:compare_local_with_transfer_learning_model}
\begin{tablenotes}
\item Notes: (1) For ``Model type", L means local model, TL means transfer learning model, G means global model. The global models are not trained on the target training set. They are only trained on the source datasets and applied to the target test set. (2) The test RMSE is calculated based on the ensemble mean prediction of the ten sub-models from ten-fold cross-validation. (3) For ``Training samples", the meaning is different for different types of models. For local and transfer learning models, we count the number of training samples used for the target model without counting those used to develop the source model; this is because large-scale pre-trained models will mostly be open-source and accessible, and users do not need to train by themselves. For the global model, we count the number of training samples for the source model as it is directly applied to the target test set without training a target model. (4) ``Training epochs" is the epochs needed until model convergence, which is represented by mean$\pm$std based on ten sub-models from ten-fold cross-validation. (5) Training effort=Training samples $\times$ Training epochs, and M stands for a million samples.
\end{tablenotes}
\end{small}
\end{threeparttable}
\end{adjustbox}
\end{table}

\section{Conclusion}
\label{sec:conclusion}
In this study, we explore developing deep learning-based solar forecasting models based on three heterogeneous datasets collected from the world. We compare the performance of models trained individually based on local datasets (local models) and models trained jointly based on the combination of multiple datasets (global models), and we further examine the potential of knowledge transfer from pre-trained solar forecasting models to a new dataset of interest (transfer learning models). 

The results suggest that the local models works best when deployed locally. Significant errors are observed for the scale of the prediction when applied offsite, however, the trend and shape of the PV/irradiance time series can be predicted well, which indicates the feature representations learned by the local models can generally be shared, while the learned regression functions are location dependent. Training on a global dataset can improve the generalization of the model, evidenced by better performance when adapting to individual locations, though one should be aware of the possible increase in training effort required. Transfer learning can help significantly when there is limited data for training, especially if a source model is pre-trained on a large and diversified source dataset. The results show that by pre-training solar forecasting models on the combined global dataset and then transferring to the local DEWA dataset, either using a warm-starting or freezing convolutional blocks strategy, it can reduce the training effort by 80\% while achieving 1\% improvement in prediction accuracy compared with the local baseline model trained with the whole dataset. Thus, we further call on the community to contribute to a large-scale global dataset with massive and diversified data for solar forecasting. Open-sourcing pre-trained models built upon such a large-scale dataset can benefit the local model development by saving significantly on training effort. Future studies will also investigate the benefit of data augmentation (e.g., image mixing, rotations, vertical/horizontal flips) and scene representation (e.g., polar coordinates, sun-centred sky images) on knowledge transfer in more diverse contexts.

\section*{Data availability}
The Stanford dataset can be accessed at \url{https://purl.stanford.edu/dj417rh1007} with detail information documented in \url{https://github.com/yuhao-nie/stanford-solar-forecasting-dataset}, and the SIRTA dataset is available upon request to \url{https://sirta.ipsl.polytechnique.fr/data_form.html}.

\section*{Acknowledgements}
The research was supported by the Dubai Electricity and Water Authority (DEWA) through their membership in the Stanford Energy Corporate Affiliates (SECA) program. This work was also sponsored by ENGIE Lab CRIGEN, EPSRC (EP/R513180/1) and the University of Cambridge. The authors acknowledge SIRTA for sharing the data used in this study and the Stanford Research Computing Center for providing the computational resources for conducting the experiments.

\bibliography{mybibfile}

\end{document}